\documentclass[10pt,twocolumn,letterpaper]{article}

\usepackage[pagenumbers]{cvpr} 

\usepackage{graphicx}
\usepackage{amsmath}
\usepackage{amssymb}
\usepackage{booktabs}

\usepackage{multirow}
\usepackage{color, colortbl}
\usepackage{subcaption}
\usepackage{tabu}
\usepackage{pifont}
\usepackage{makecell}
\usepackage{paralist}
\usepackage{threeparttable}
\usepackage{enumitem}

\newcommand{\xmark}{\ding{55}}%

\usepackage[rgb,dvipsnames]{xcolor}

\newcommand{\xhdr}[1]{\vspace{4pt} \noindent {\textbf{#1}}}

\usepackage[pagebackref,breaklinks,colorlinks]{hyperref}

\usepackage[capitalize]{cleveref}
\crefname{section}{Sec.}{Secs.}
\Crefname{section}{Section}{Sections}
\Crefname{table}{Table}{Tables}
\crefname{table}{Tab.}{Tabs.}

\def\cvprPaperID{****} 
\def\confName{****}
\def\confYear{****}

\def\Modelname{\textsc{VindLU}}
	
\definecolor{Gray}{gray}{0.5}
\definecolor{LGray}{gray}{0.9}

\usepackage{textcomp}  
\usepackage{scalerel}  

\def\currynew{\scalerel*{\includegraphics{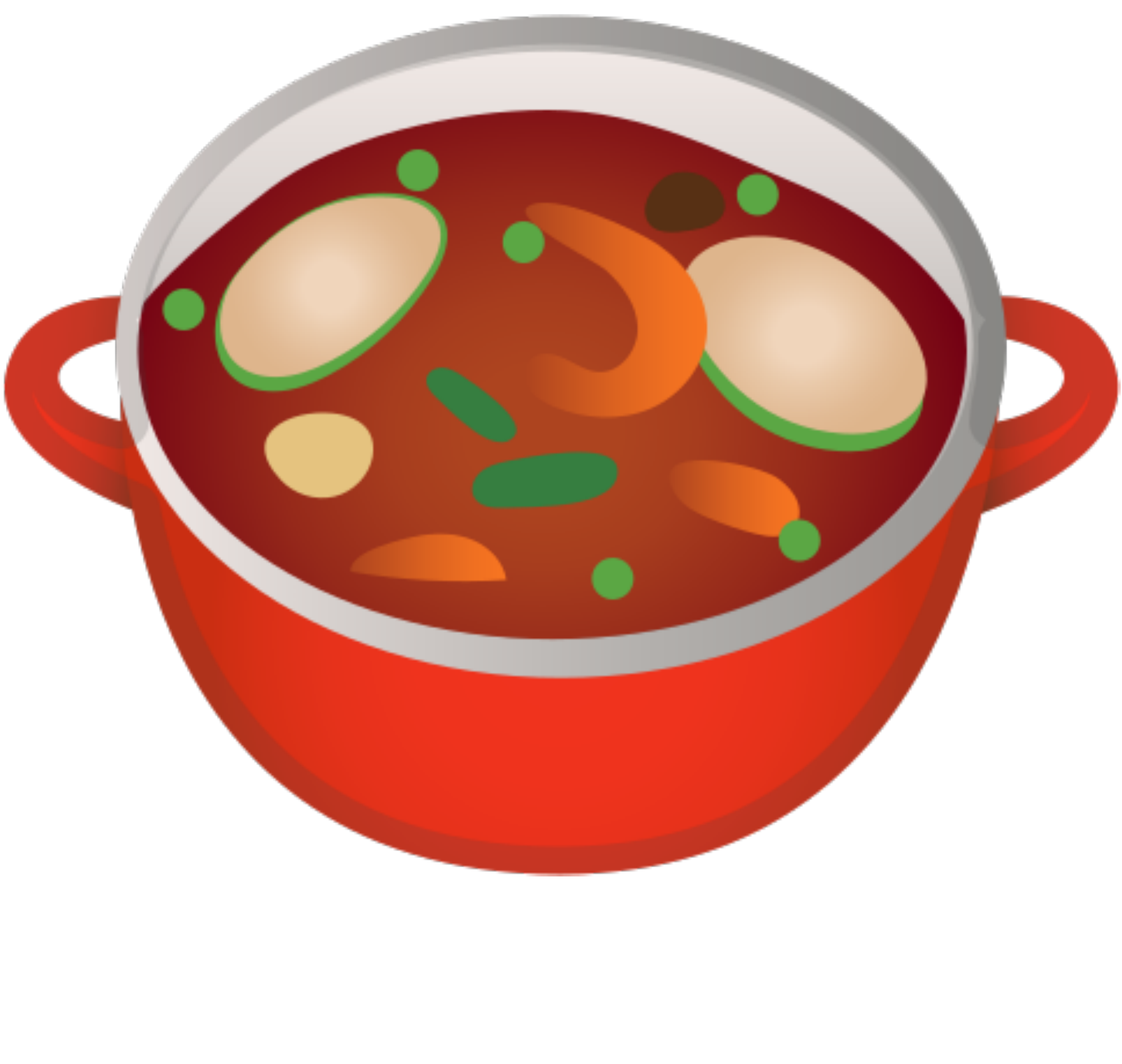}}{\textrm{\textbigcircle}}}

\begin{document}

\title{\textsc{VindLU} \currynew: A Recipe for Effective Video-and-Language Pretraining }

\author{Feng Cheng$^1$ \quad Xizi Wang$^2$  \quad Jie Lei$^1$  \quad David Crandall$^2$  \quad Mohit Bansal$^1$  \quad Gedas Bertasius$^1$ \\
 $^1$UNC Chapel Hill \quad $^2$Indiana University \\
{\tt\small \{fengchan,jielei,mbansal,gedas\}@cs.unc.edu \quad \{xiziwang, djcran\}@iu.edu}
}

\maketitle

\begin{abstract}
The last several years have witnessed remarkable progress in video-and-language (VidL) understanding. However, most modern VidL approaches use complex and specialized model architectures and sophisticated pretraining protocols, making the reproducibility, analysis and comparisons of these frameworks difficult. Hence, instead of proposing yet another new VidL model, this paper conducts a thorough empirical study demystifying the most important factors in the VidL model design. Among the factors that we investigate are (i) the spatiotemporal architecture design, (ii) the multimodal fusion schemes, (iii) the pretraining objectives, (iv) the choice of pretraining data, (v) pretraining and finetuning protocols, and (vi) dataset and model scaling. Our empirical study reveals that the most important design factors include: temporal modeling, video-to-text multimodal fusion, masked modeling objectives, and joint training on images and videos. Using these empirical insights, we then develop a step-by-step recipe, dubbed \Modelname, for effective VidL pretraining. Our final model trained using our recipe achieves comparable or better than state-of-the-art results on several VidL tasks without relying on external CLIP pretraining. In particular, on the text-to-video retrieval task, our approach obtains 61.2\% on DiDeMo, and 55.0\% on ActivityNet, outperforming current SOTA by 7.8\% and 6.1\% respectively. Furthermore, our model also obtains state-of-the-art video question-answering results on ActivityNet-QA, MSRVTT-QA, MSRVTT-MC and TVQA. Our code and pretrained models are publicly available at: \url{https://github.com/klauscc/VindLU}.
\end{abstract}

\section{Introduction}
\label{sec:intro}

\begin{figure}[t] 
\centering
\includegraphics[width=0.46\textwidth]{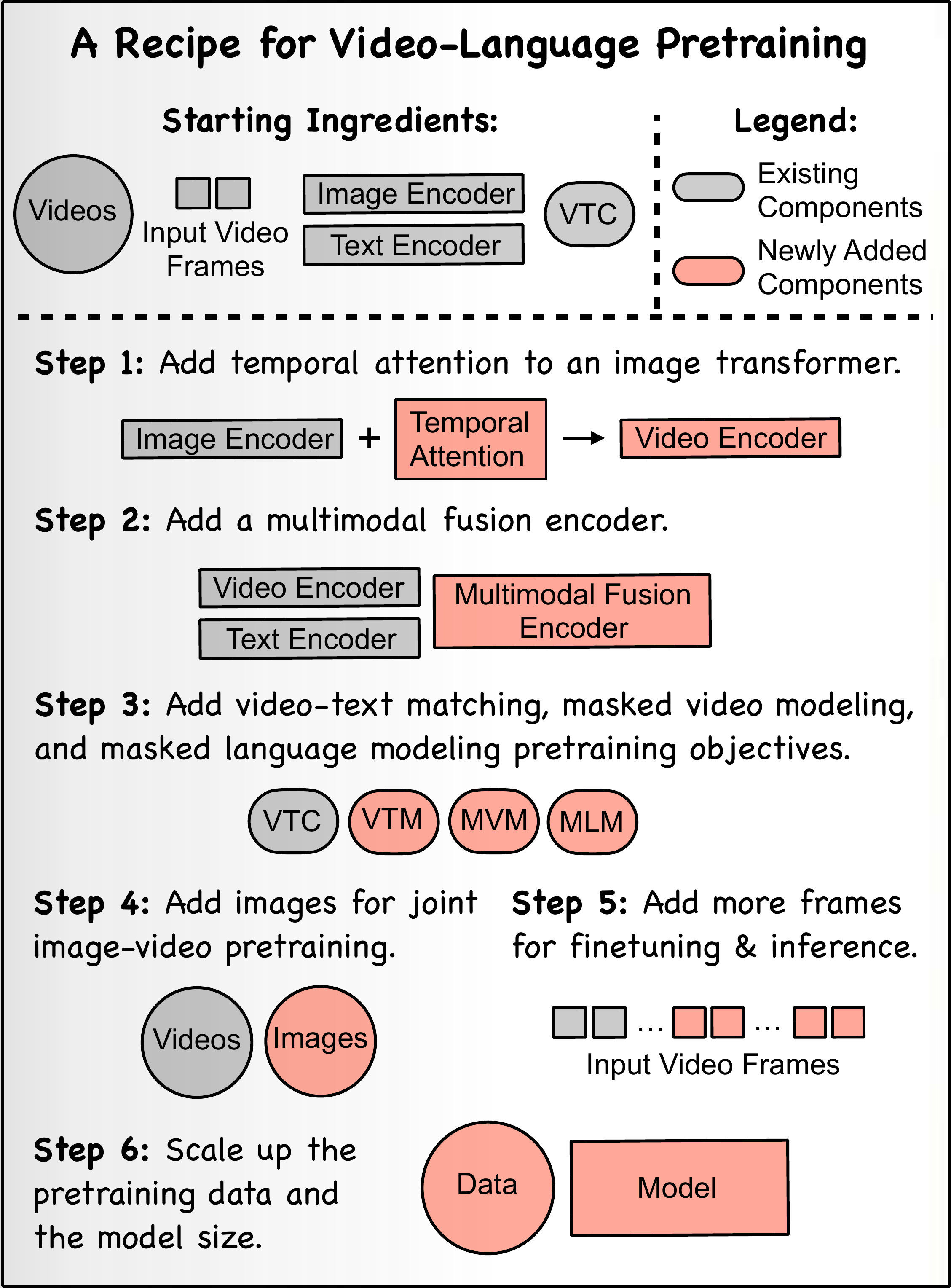}
\caption{
We present a recipe for effective video-language pretraining. Our recipe starts with image and text transformer encoders trained on video-text pairs using a contrastive objective (VTC).  We then progressively add more components to our framework while also studying the importance of each component along the way. Our final recipe includes the steps for (1) adding temporal attention, (2) injecting a multimodal fusion encoder, (3) incorporating masked modeling pretraining objectives, (4) jointly training on images and videos, (5) using more frames during fine-tuning and inference, and lastly, (6) scaling up the data and the model.
}
\label{fig:teaser}
\end{figure}

\begin{table*}[htpb]
    \small
    \centering
    \begin{threeparttable}
    \setlength{\tabcolsep}{2pt}
    \begin{tabular}{lccccrcccc}
    \toprule
    \multirow{2}{*}{\textbf{Method}}& \multicolumn{3}{c}{\textbf{Model Design}} &\multicolumn{3}{c}{\textbf{Pretraining Data}} & \multicolumn{3}{c}{\textbf{\#Frames}} \\
    \cmidrule(lr){2-4} \cmidrule(lr){5-7} \cmidrule{8-10}
     & \makecell{Temporal \\ Modeling} & \makecell{Multimodal \\ Fusion} & \makecell{Pretraining \\ Objectives} & Dataset & Size & Modality & PT & FT & Eval \\
    \midrule
    UniVL\cite{luo2020univl}& Joint Att.\cite{gberta_2021_ICML} & 2-layer TR & VTC+VTM+MLM+MFM+LM  &HT & 136M & V & 48 & 48 & 48 \\
    VideoCLIP\cite{xu2021videoclip}  & 1D-Conv+TR & \xmark & VTC & HT & 136M & V & 32 & 32 & 32 \\
    ClipBert\cite{lei2021less} & Mean Pooling & BERT & MLM+VTM & COCO+VG & 0.2M & I & 1 & 16 & 16 \\
    Frozen\cite{bain2021frozen} & Temp. Attn\cite{gberta_2021_ICML} & \xmark & ITC & C5M & 5M & I+V &  $1\to4$ & 4& 4\\
    MERLOT~\cite{zellers2021merlot} & Joint Attn & RoBERTa & VTC+MLM+FOM & YT & 180M & V & 16 & 16&  16\\
    VIOLET~\cite{fu2021violet} & Window Attn~\cite{liu2022video} & BERT & \makecell{VTC+VTM+MLM+MVM}  & YT+C5M& 185M & I+V & 4 & 5 & 5\\
    MV-GPT~\cite{seo2022end} & Joint Attn & 2-layer TR & MLM+LM & HT & 136M & V & - & - & - \\
    ALL-in-one~\cite{wang2022all} & Token Rolling~\cite{wang2022all} & ViT & VTC+VTM+MLM &HT+W2 & 172M & V & 3 & 3& 9\\
    Singularity~\cite{lei2022revealing} & Late Temp. Attn & 3-layer TR & VTC+VTM+MLM & C17M & 17M & I+V & $1\to4$ & 4& 12 \\
    LAVENDER~\cite{li2022lavender} & Window Attn~\cite{liu2022video} & BERT & MLM & C17M+IN & 30M & I+V & 4 & 5 & 5 \\
    OmniVL~\cite{wang2022omnivl} & Temp. Attn & $2\times$BERT & VTC+VTM+LM & C17M & 17M & I+V & $1\to8$ & 8 & 8 \\
    ATP~\cite{buch2022revisiting} & \xmark & \xmark & VTC &CLIP & 400M & I & 1 & 16 & 16  \\
    CLIP4Clip~\cite{luo2022clip4clip} & Late TR & \xmark & VTC & CLIP & 400M & I & 1 & 12& 12 \\
    ECLIPSE~\cite{lin2022eclipse} & Late TR & \xmark & VTC & CLIP & 400M & I+A & 1 & 32 & 32 \\
    CLIP2TV~\cite{gao2021clip2tv} & CLIP & 4-layer TR & VTC+VTM & CLIP & 400M & I & 1 & 12 & 12\\
    CLIP-Hitchhiker~\cite{bain2022clip} & Late Attn & \xmark & VTC & CLIP & 400M & I & 1 & 16 & 120 \\
    CLIP-ViP~\cite{xue2022clip} & Prompt Attn~\cite{xue2022clip} & \xmark & VTC & CLIP & 500M & I+V & $1\to12$ & 12 & 12 \\
    \bottomrule
    \end{tabular}
    \begin{tablenotes}\footnotesize
    \item[] \textbf{TR}: Transformer; \textbf{Late}: Late fusion; \textbf{Attn}: Attention. \textbf{V}: Video; \textbf{I}: Image; \textbf{A}: Audio; $1 \to 4$: $1$ frame for stage-1 training and $4$ frames for stage-2. \\
    \textbf{VTC}: Video-text contrastive; \textbf{VTM}: Video-text matching; \textbf{MLM}: Masked language modeling; \textbf{MFM}: Masked frame modeling; \textbf{LM}: Language modeling.
    \textbf{HT}: HowTo100M~\cite{miech2019howto100m}; \textbf{C5M, C17M}: see supplementary; \textbf{YT}: YT-Temporal~\cite{zellers2021merlot}; \textbf{W2}: WebVid-2M~\cite{bain2021frozen}; \textbf{COCO}:~\cite{lin2014microsoft}, \textbf{VG}: Visual Genome~\cite{krishna2017visual}; \textbf{IN}: An internal dataset.
    \end{tablenotes}
    \end{threeparttable}
    \caption{An overview of the existing VidL methods.
    Significant differences exist among these methods, making it challenging to reproduce, analyze and compare these methods.
    This motivates us to answer the question ``What are the key steps to build a highly performant VidL framework" by investigating various components in the VidL framework design.
    }
    \label{tab:existing_work}
\end{table*}

Fueled by the growing availability of video-and-text data~\cite{miech2019howto100m, bain2021frozen,chen2015microsoft,krishna2017visual,ordonez2011im2text,sharma2018conceptual,changpinyo2021conceptual} and advances in the Transformer model design~\cite{vaswani2017attention,dosovitskiy2020image}, the last few years have witnessed incredible progress in video-and-language (VidL) understanding~\cite{zhu2020actbert,xu2021videoclip,li2020hero,lei2021less,zellers2021merlot,luo2022clip4clip}. Since the initial transformer-based models for VidL, such as ClipBERT~\cite{lei2021less}, the text-to-video retrieval accuracy has improved from $22.0\%, 22.4\%$, and $21.3\%$ on MSR-VTT~\cite{xu2016msr}, DiDeMo~\cite{anne2017localizing}, and ActivityNet~\cite{krishna2017dense} to $>45\%$ R@1 accuracy on all three of these datasets, thus, marking an extraordinary relative improvement of more than $100\%$ in less than $2$ years. 

At the same time, the model architectures and pretraining/finetuning protocols used by modern VidL approaches have become significantly more complex and specialized over the last several years. As a result, it is increasingly difficult to reproduce, analyze and compare most recent VidL frameworks. For example, several recent approaches~\cite{lei2022revealing,li2022lavender,xue2022clip} propose new architectures, new initialization strategies, pretraining objectives, pretraining datasets, and optimization protocols. Due to the large computational cost of ablating all these factors, it is difficult to understand which components are critical to the success of the proposed frameworks. Similarly, the key success factors of many other recent VidL approaches~\cite{wang2022omnivl,fu2021violet, li2022lavender,buch2022revisiting} are also often obfuscated, which hinder future research.

In Table~\ref{tab:existing_work}, we illustrate the complexity of modern VidL frameworks by dissecting them along multiple dimensions, including temporal modeling schemes, multimodal fusion modules, pretraining objectives, the source of the pretraining data, and the number of frames for pretraining, finetuning and inference.
Based on this analysis, we observe that there exist significant differences among these VidL methods. Unfortunately, it's not clear which differences are important for the overall VidL performance and which are not.

The recent METER~\cite{dou2022empirical} work studies a subset of these components in the context of image-language modeling. However, their analysis is limited to images and, thus,  ignores various aspects related to video modeling, such as spatiotemporal architecture design, video pretraining objectives, video pretraining data, and video-specific finetuning/evaluation protocols such as the number of frames.
As we will show in our experimental section, many of the findings presented in the image-based studies~\cite{dou2022empirical} do not hold for video.
Beyond image-based analysis, we note that the concurrent work in~\cite{fu2022empirical} conducts an empirical study of VidL transformers. However, unlike our work, which covers a broad range of VidL design factors, their analysis is focused predominantly on masked visual modeling objectives, which we also study in this work.

Our main objective in this work is to answer the question ``What are the key steps needed to build a highly performant VidL framework?" To do this, we conduct a thorough empirical study that demystifies the importance of various VidL design choices and ultimately leads to a VidL framework that achieves state-of-the-art results on various VidL benchmarks. Using our empirical insights, we then develop a step-by-step recipe for effective VidL pretraining. Our recipe, dubbed \Modelname~(VIdeo aND Language Understanding), starts from a standard Vision Transformer (ViT)~\cite{dosovitskiy2020image} and uses a simple progressive expansion scheme where at each step, we investigate a particular aspect of VidL framework design (e.g., architecture, pretraining objective, pretraining data, etc.), and choose the best performing option. In particular, we study the following VidL design components: (i) the spatiotemporal architecture design, (ii) the multimodal fusion schemes, (iii) the pretraining objectives, (iv) the source of the pretraining data, (v) finetuning/inference protocols, and (vi) scaling of the data and model. We present our recipe in Fig.~\ref{fig:teaser}.

\begin{figure}
    \centering
    \includegraphics[width=\linewidth]{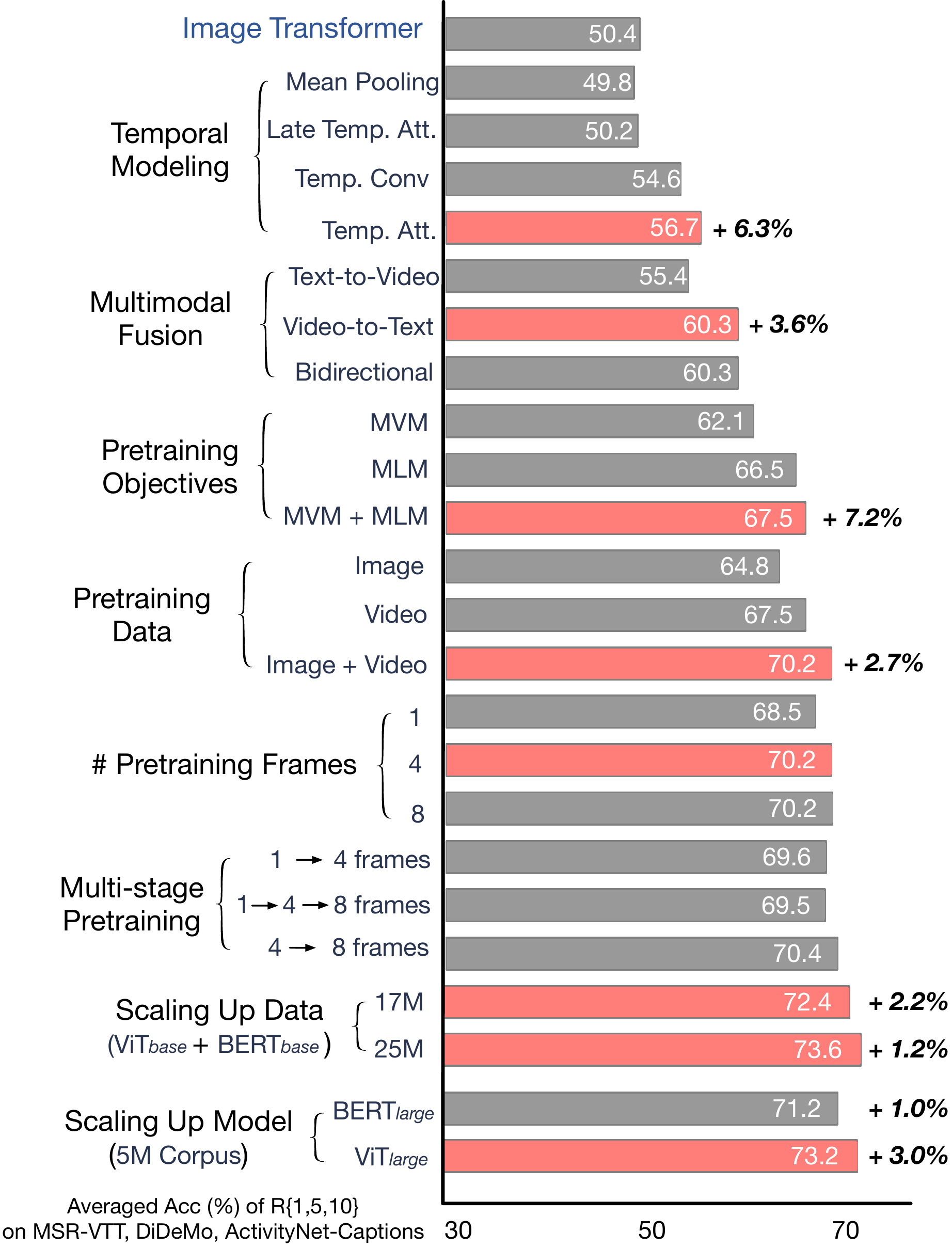}
    \caption{We progressively expand an image transformer baseline (e.g., ViT) to a performant video-and-language (VidL) model. We do so by investigating the importance of many VidL design choices such as (i) temporal modeling, (ii) multimodal fusion modules, (iii) pretraining objectives, (iv) the source of the pretraining data, (v) the number of pre-training frames, (vi) multi-stage pretraining, and (vii) scaling of the data and model. Each bar depicts an average text-to-video retrieval Recall@{1,5,10} accuracy across MSR-VTT~\cite{xu2016msr}, DiDeMo~\cite{xu2016msr}, ActivityNet~\cite{krishna2017dense}. 
    The red bars denote the best-performing design choice in each subgroup. Our final VidL framework, dubbed \Modelname, outperforms our initial image Transformer baseline by \textbf{23.2\%}. The figure was inspired by~\cite{liu2022convnet}.
    }
    \label{fig:expansion}
\end{figure}

The key findings of our empirical study include:
{\setlength{\parindent}{0em}
\begin{itemize}[nosep,leftmargin=1em,labelwidth=*,align=left]
  \item Contrary to the conclusions of several prior works~\cite{lei2022revealing,buch2022revisiting} that a single frame is sufficient for VidL modeling, we discover that temporal modeling using multiple frames leads to a significant improvement over the spatial-only baselines (\textbf{+6\%} averaged video retrieval accuracy on MSR-VTT, DiDeMo, and ActivityNet).
  \item Multimodal fusion module that incorporates video features into text is critical for good VidL performance (\textbf{+3.6\%}). Conversely, we find that adding text features to the video representation is not useful.
  \item Masked language modeling objective significantly improves performance (\textbf{+6.2\%}). However, to obtain such gains, a BERT-like language model pretrained on this objective is needed for initialization. Masked video modeling objective brings an additional \textbf{+1\%} improvement. 
  \item Pretraining jointly on images and videos is beneficial (\textbf{+2.7\%}). Also, contrary to prior methods~\cite{bain2021frozen,wang2022omnivl}, we find  multi-stage training unnecessary. 
  \item Pretraining with a small number of frames (e.g., 4) is sufficient and it can significantly reduce the computational cost of large-scale pretraining. Pretraining with more frames does not lead to a substantial performance boost.
  \item Compared to many recent CLIP-based~\cite{radford2021learning} VidL approaches~\cite{luo2022clip4clip,xue2022clip,bain2022clip}, our recipe achieves comparable or even better performance with $\bf{20\times}$ less pretraining data.
\end{itemize}
}

\noindent Our final model, trained using our \Modelname~recipe, achieves state-of-the-art results on several VidL benchmarks. Specifically, on the video retrieval task, our method achieves 46.5\%, 61.2\%, 55.0\% R@1 accuracy on MSR-VTT, DiDeMo, and ActivityNet, outperforming the state-of-the-art by \textbf{7.8\%} and \textbf{6.1\%} on the latter two datasets. Also, our approach obtains state-of-the-art video question-answering results on ActivityNet-QA, MSRVTT-QA, MSRVTT-MC and TVQA, where we achieve top-1 accuracy of 44.7\%, 44.6\%, 95.5\%, and 79.0\% respectively.

We want to make it clear that, in this paper, we do not claim technical novelty behind any of the individual design choices (i.e., different subsets of these design choices were already used by prior VidL methods as shown in Table~\ref{tab:existing_work}).
Instead, our main contribution, which we believe might be equally if not more important than proposing yet another specialized or obfuscated VidL model, is to investigate these components collectively and validate their importance. We also do not claim superiority over previous methods (despite better results). Due to the implementation complexities of each method, fair and complete comparisons are difficult and not our intent. Instead, we hope that our recipe for building an effective VidL framework will provide useful insights for future research on VidL understanding. To enable the VidL community to build on our work, we release our code and pretrained models.

\section{Related Work}

\xhdr{Image-and-Language Pretraining.}
Recent years have witnessed remarkable progress in image-and-language pretraining~\cite{tan2019lxmert,lu2019vilbert,zhou2020unified,chen2020uniter,zhang2021vinvl,yang2021causal,radford2021learning,kim2021vilt,yuan2021florence,wang2021ufo,zhai2022lit,yang2022unified,singh2022flava,hu2022scaling,wang2022vlmixer,zeng2021multi,li2022blip,byun2022grit,wang2022image}. However, most modern methods such as ViLBERT~\cite{lu2019vilbert}, UNITER~\cite{chen2020uniter}, CoCa~\cite{yu2022coca}, LEMON~\cite{hu2022scaling}, BEiT-3~\cite{wang2022image} typically employ complex transformer-based architectures and pretraining objectives. As a result, it is difficult to decipher which components are critical for good performance. A recent empirical study on image-language modeling METER~\cite{dou2022empirical} studies a variety of components, including the choice of a vision encoder, multimodal fusion schemes, and pretraining objectives. However, since their analysis is done exclusively on images, it's unclear whether these findings generalize to video. The analysis of METER also ignores many video-specific design choices such as temporal modeling schemes, video pretraining objectives and data, and video-specific finetuning/inference protocols. In comparison, our work thoroughly studies all of these components, the result of which is a detailed step-by-step recipe for effective video-language pretraining.

\xhdr{Video-and-Language Pretraining.}
In recent years, the large-scale VidL pretraining~\cite{buch2022revisiting,fu2021violet,li2022lavender,wang2022omnivl,lei2021less,wang2022object} has shown strong transfer learning ability to downstream VidL tasks such as text-to-video retrieval~\cite{xu2016msr,anne2017localizing,krishna2017dense,lei2021less,liu2019use,luo2022clip4clip,yu2018joint}, video question answering~\cite{yu2018joint,xu2017video,yu2019activitynet}, video captioning ~\cite{krishna2017dense,iashin2020multi,wang2018reconstruction,zolfaghari2018eco,sun2019videobert}, etc.
Several methods~\cite{luo2022clip4clip,gao2021clip2tv,xue2022clip,bain2022clip} achieve impressive results by building on the popular image-language pretrained model CLIP~\cite{radford2021learning}. Additionally, several recent approaches~\cite{lei2022revealing, li2022lavender,wang2022omnivl} propose more sophisticated VidL frameworks to achieve comparable performance as CLIP-based methods without large-scale CLIP pretraining. However, with the impressive results, these methods also require more complex architectures and specialized video pretraining protocols (as shown in Table~\ref{tab:existing_work}).
The complexity of these frameworks and the large computational cost of VidL pretraining makes it challenging to decipher which VidL framework components are truly needed for good performance. Moreover, unlike in the image-language domain, there are few empirical studies investigating various VidL design components collectively. For instance, the concurrent work of Fu~\cite{fu2022empirical} only studies masked video modeling pretraining objectives and is based on a slightly older VIOLET~\cite{fu2021violet} method. Furthermore, the recent works~\cite{lei2022revealing, buch2022revisiting} focus predominantly on spatial biases in modern VidL benchmarks.
In contrast to these prior approaches, our work aims to investigate the importance of a broad range of factors in VidL framework design. We then use our empirical insights to provide a detailed step-by-step recipe for effective VidL pretraining. 

\section{A Recipe for Video-Language Pretraining}
\label{sec:method}

In this section, we describe our recipe for video-and-language (VidL) pretraining. We begin with a standard image transformer (e.g., ViT~\cite{dosovitskiy2020image}) and progressively expand it to a model that achieves state-of-the-art results on various VidL datasets and tasks. At each step of our recipe, we study how various design choices affect VidL performance. In particular, we are interested in answering the following questions about the VidL pretraining design:

{\setlength{\parindent}{0em}
\begin{itemize}[nosep,leftmargin=1em,labelwidth=*,align=left]
    \item Does a VidL model benefit from a temporal modeling capability, especially considering that most VidL benchmarks are spatially biased as demonstrated by several prior methods~\cite{buch2022revisiting,lei2022revealing}? If so, what is the best mechanism for temporal modeling?
    \item What is the most effective way to do multimodal fusion? Some prior approaches~\cite{fu2021violet,wang2022all,li2022lavender} use bidirectional whereas others~\cite{lei2022revealing,wang2022omnivl} employ unidirectional (e.g., text-to-video or video-to-text) multimodal fusion modules. Which of these fusion schemes works the best? 
    \item Which pretraining objectives are most useful for VidL representation learning? Previous methods adopt many pretraining objectives including video-text contrastive (VTC)\cite{li2022align}, video-text matching (VTM)\cite{li2022align,li2020hero,luo2020univl}, masked-language-modeling (MLM)\cite{devlin2018bert}, and masked-video-modeling (MVM)\cite{tong2022videomae}. How important are each of these objectives? Are they complementary to each other?
    \item What pretraining data is most useful for training VidL models? Should we train VidL models only on the video data or jointly on images and videos? If so, how do we do this effectively? Prior works~\cite{bain2021frozen, wang2022omnivl,wang2022all} propose a variety of different pretraining protocols (e.g., a single-frame training, curriculum learning, joint multi-frame pretraining, etc.). Which of these is the most effective?
    \item How many frames are needed for pretraining, fine-tuning, and inference? Several recent approaches~\cite{lei2022revealing,buch2022revisiting} claimed that single frame pretraining is sufficient while others~\cite{wang2022omnivl,xue2022clip} proposed to pretrain their models with 8 or even more frames. Furthermore, should we finetune and test the pretrained VidL models using the same number of frames as during pretraining? Is it helpful to use more frames during fine-tuning and inference?
\end{itemize}
}

\noindent Motivated by these questions, we next present our recipe while also studying these questions in more detail. 

\subsection*{Step 0: Starting Ingredients}

\xhdr{Image Transformer Baseline.} We start with a standard ViT-B/16~\cite{dosovitskiy2020image} transformer trained on single frames of the WebVid-2M dataset\cite{bain2021frozen}. For text encoder, we use BERT~\cite{devlin2018bert} throughout all of our experiments. Formally, given the paired video and text input $(v, t)$, the image transformer randomly selects a single frame from the video as input to extract the video embeddings. A text encoder encodes the text $t$ to extract the text embeddings. We then use a video-text contrastive (VTC) loss to maximize the agreement between the paired video and text embeddings as in~\cite{bain2021frozen,radford2021learning}. Following ~\cite{lei2022revealing}, we use BEiT \cite{bao2021beit} initialization for our image transformer, whereas the text encoder is initialized with $\text{BERT}_{base}$.

\xhdr{Experimental Setup.} As our initial pretraining data, we use WebVid-2M\cite{bain2021frozen} unless noted otherwise. Afterward, we finetune and evaluate our pretrained model on the three popular text-to-video retrieval datasets: MSR-VTT\cite{xu2016msr}, DiDeMo\cite{anne2017localizing}, and ActivityNet-Captions\cite{krishna2017dense}, which include both short and long videos. As our evaluation metric, we report the averaged Top-1, Top-5, and Top-10 text-to-video retrieval accuracy across these three datasets. As shown in the Fig.~\ref{fig:expansion}, our Image Transformer baseline achieves an average accuracy of \textbf{50.4\%}. 

Over the next several subsections, we progressively expand this baseline by adding more components of increasing complexity. In particular, we start by incorporating (i) temporal modeling blocks, (ii) a multimodal fusion encoder, and (iii) additional pretraining objectives. Afterward, we investigate the choice for the (iv) pretraining data, (v) finetuning and inference protocols, and (vi) dataset and model scaling schemes. We would like to note that due to the large computational cost, we cannot ablate the order of the steps in our recipe. Thus, the order of the steps is primarily determined by the computational cost (i.e., the steps that can be implemented most efficiently are studied first then, moving to the more computationally costly steps).  

\subsection*{Step 1: Temporal Modeling}
\label{step:temporal_modeling}

In the first step of our recipe, we extend our initial image transformer to video via a temporal modeling mechanism, which enables training our model on multiple frames. Such a temporal modeling mechanism would enable training our model on multiple frames for more robust VidL spatiotemporal representation learning. For compactness, in this part of our empirical study, we include the analysis of the four commonly used temporal modeling schemes. More temporal modeling baselines can be found in Appx.~\ref{appendix:temporal_modeling}.

{\setlength{\parindent}{0em}
\begin{itemize}[nosep,leftmargin=1em,labelwidth=*,align=left]
\item \textbf{Mean Pooling (MP).} In this variant, the visual encoder processes input frames independently and averages their frame-wise scores for the video-level score as in~\cite{luo2022clip4clip}.

\item  \textbf{Late Temporal Attention (L-TA).} Following~\cite{lei2022revealing,neimark2021video,luo2022clip4clip} we use a late temporal modeling scheme by attaching 2 Transformer layers to an image encoder, which then aggregates temporal information across all input frames.
\item \textbf{Temporal Convolution (TC).} Many previous methods~\cite{xie2018rethinking,pan2022st,feichtenhofer2020x3d} used 3D convolutions for temporal modeling.
To validate its effectiveness, we inject a TC block, consisting of a linear down-projection layer with hidden size 384, a depth-wise $3\times1\times1$ convolution as in~\cite{Tran_2019_ICCV}, a ReLU activation, and a linear up-projection layer, before the spatial attention to each Transformer Layer.

\item  \textbf{Temporal Attention (TA).} Inspired by TimeSformer~\cite{gberta_2021_ICML}, we experiment with divided space-time attention, which we insert before spatial attention as in~\cite{gberta_2021_ICML}.
\end{itemize}
}

\noindent As shown in the upper part of Fig.~\ref{fig:expansion} and the Table below, the temporal modeling capability is critical for good VidL performance. This is indicated by a $\bf+6.3\%$ accuracy boost of our temporal attention variant (TA) over the spatial-only baseline. We also observe that late temporal modeling (L-TA) has nearly no effect. We conjecture that this is due to the limited temporal modeling capacity (\ie, only two layers) and the lack of temporal fusion in the early layers. Lastly, our results suggest that TA outperforms TC by \textbf{2.1\%}, which might indicate that long-range temporal attention is more useful than local 3D convolutions.  

\begin{table}[h]
    \centering
    \small
    \begin{tabular}{l|cccc}
         & Mean Pooling & L-TA & TC & TA \\
        \hline
        acc.(\%) & 49.8 & 50.2 & 54.6 & \textbf{56.7}
    \end{tabular}
\end{table}

Interestingly, we note that our findings are contrary to the conclusions of several recent methods~\cite{buch2022revisiting,lei2022revealing} claiming that temporal modeling is not needed for many VidL tasks. Upon experimenting with the publicly released models of~\cite{lei2022revealing}, we found that the temporal variants of their approach performed consistently better than the spatial-only variants, further strengthening our conclusions. We conjecture that even on the spatially-biased datasets, temporal modeling might be useful for resolving spatial ambiguities caused by appearance variations across different frames.

\textit{Takeaway \#1: For all subsequent experiments, we adopt Temporal Attention (TA) as our temporal modeling mechanism and pretrain our model with 4-frame inputs unless otherwise noted. 
}

\subsection*{Step 2: Multimodal Fusion Encoder}
\label{step:multimodal_fusion}

Building on the model from Step 1 (Fig.~\ref{fig:multimodal}a), we next analyze the role of multimodal fusion modules.  The purpose of the multimodal fusion encoder is to fuse multimodal cues from video and language for a more discriminative VidL feature representation. As shown in Fig.~\ref{fig:multimodal}, we experiment with several variants of multi-modal fusion encoders:

\begin{figure*}
    \centering
    \includegraphics[width=\linewidth]{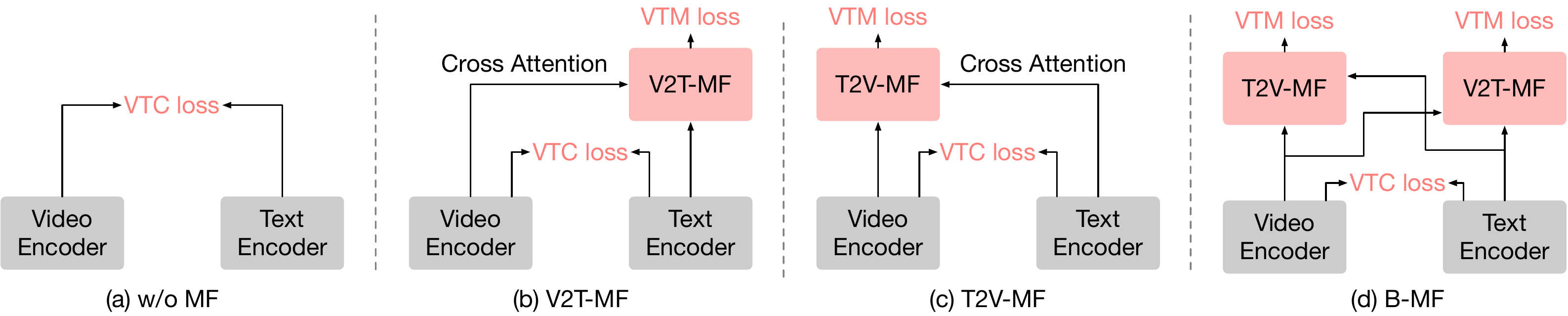}
    \caption{A high-level illustration of different multimodal fusion schemes: (a) a baseline  that does not use any multimodal fusion (w/o MF),
    (b) video-to-text multimodal fusion (V2T-MF), (c) text-to-video multimodal fusion (T2V-MF) and (d) bidirectional multimodal fusion (B-MF).
    The video-text matching (VTM) loss is attached to the multimodal fusion encoder whereas video-text contrastive (VTC) loss is added to the video and text encoders. The entire framework is trained end-to-end using VTM and VTC losses.
    }
    \label{fig:multimodal}
\end{figure*}

{\setlength{\parindent}{0em}
\begin{itemize}[nosep,leftmargin=1em,labelwidth=*,align=left]
 \item \textbf{Video-to-Text Multimodal Fusion (V2T-MF).} As illustrated in Fig.~\ref{fig:multimodal}b, V2T-MF injects relevant video cues into the textual features using Cross-Attention. For a fair comparison with previous baselines~\cite{bain2021frozen,lei2022revealing}, we do not add any extra layers but instead re-purpose the last $m$ layer of our text encoder for V2T fusion. Specifically, a cross-attention operation is inserted into each of the $m$ last layers in the text encoder between Self-Attention and MLP. This scheme was also previously used by~\cite{lei2022revealing,wang2022omnivl}.

\item \textbf{Text-to-Video Multimodal Fusion (T2V-MF).} Similar to V2T-MF, we build T2V-MF (Fig.~\ref{fig:multimodal}c) by re-purposing the last $m$ layers of the vision encoder and utilizing cross-attention to incorporate text cues into the video features.

\item \textbf{Bidirectional Multimodal Fusion (B-MF).}
Instead of using unidirectional multimodal fusion modules, several prior approaches~\cite{fu2021violet,li2022lavender,wang2022all,zellers2021merlot} concatenate the visual features and text features and then feed them jointly to a subsequent $m$-layer multimodal Transformer. However, this is often infeasible in the video domain due to a large number of input frames and hence, large computational cost. Thus, instead, we implement B-MF (Fig.~\ref{fig:multimodal}d) by combining the T2V-MF and V2T-MF, which reduces the space and time complexity from $O((L + S)^2)$ to $O(L \times S)$ where $L, S$ is the number of video tokens and text tokens respectively.
\end{itemize}
}

\noindent To train each multimodal fusion encoder variant, we add the video-text matching (VTM) loss objective (described in the Sec.~\ref{subsubsec:step3_loss}) as was done in several prior approaches~\cite{wang2022omnivl,fu2021violet,wang2022all}.
In the table below and Figure~\ref{fig:expansion}, we present our analysis. Based on these results, we report that the V2T-MF scheme performs the best (i.e., \textbf{+3.6\%} improvement). Surprisingly, we observe that the reverse, T2V-MF scheme, leads to  a substantially decreased performance (\textbf{-1.3\%}). We conjecture that predicting the matching video-text pairs using a pretrained language rather than a visual representation is easier. Lastly, the bidirectional fusion scheme, B-MF, yields no improvement compared to V2T-MF. We conjecture that this happens because of the poor performance in the T2V-MF branch.

\begin{table}[h]
    \centering
    \small
    \begin{tabular}{l|cccc}
         & w/o. MF & T2V-MF & V2T-MF & B-MF \\
        \hline
        acc.(\%) & 56.7 & 55.4 & \textbf{60.3} & \textbf{60.3}
    \end{tabular}
\end{table}

\textit{Takeaway \#2: For our remaining experiments, we use V2T-MF as our multimodal fusion encoder.}

\subsection*{Step 3: Pretraining Objectives}\label{subsubsec:step3_loss}

Building on the model from Step 2, we next study the following pretraining objectives: 

{\setlength{\parindent}{0em}
\begin{itemize}[nosep,leftmargin=1em,labelwidth=*,align=left]

\item \textbf{Visual-Text Contrastive Learning (VTC)}.
VTC aims to learn independent representations for video and text by maximizing the agreement between positive \emph{(visual, text)} pairs while minimizing the agreement between negative pairs. Note that this objective is already used in previous steps, and thus, not included in Figure~\ref{fig:expansion}.

\item \textbf{Visual-Text Matching (VTM)}.
VTM objective is implemented as a standard cross-entropy loss that encourages a VidL model to produce binary predictions indicating whether a given video-text pair matches. Following~\cite{li2021align}, we attach this loss to our multimodal fusion encoder and use hard negative mining during training as in~\cite{lei2022revealing}.  The VTM objective is already used in Step 2 (i.e., the multimodal fusion step) and thus, not included in Figure~\ref{fig:expansion}.

\item \textbf{Masked Language Modeling (MLM).} MLM objective aims to predict the masked words by leveraging information from both visual and textual features. To implement this pretraining objective, We mask 50\% text tokens using the same masking strategy as in BERT~\cite{zhu2020actbert} and attach a linear layer on top of our text-to-video multimodal fusion encoder (T2V-MF) to predict the masked words. See Appx.~\ref{exp:ablation:mlm} for masking ratio ablation.

\item \textbf{Masked Video Modeling (MVM)}. Just like MLM, the MVM objective aims to recover the masked tokens but in the video modality. 
This pretraining objective has been recently adopted by many self-supervised learning video methods~\cite{he2022masked,liu2022video,feichtenhofer2022masked,tong2022videomae}. To implement MVM, we apply a linear layer on top of the vision encoder and predict the masked tokens. Following~\cite{bao2021beit,he2022masked}, we randomly mask 75\% tokens and predict the masked tokens quantized using discrete variational autoencoder (dVAE)~\cite{ramesh2021zero}.
\end{itemize}
}

\noindent Based on the results in the Table below and Fig.~\ref{fig:expansion}, we observe that the MLM pretraining objective leads to a substantial boost in performance (\textbf{+6.2\%}). Furthermore, we note that adding MVM loss further improves the accuracy by \textbf{1\%}. Interestingly, our finding is contrary to the conclusions in the image-based analysis of METER~\cite{fu2022empirical}, which finds that MVM objective applied to images substantially degrades the performance.
We hypothesize that videos are more redundant compared to images, which might make the optimization easier, thus, leading to a performance boost. However, adding the MVM objective slows the training by about 40\% (due to additional forward and backward passes). Thus, to speed up the training, we don't use MVM loss in our remaining experiments.

\begin{table}[h]
    \centering
    \small
    \begin{tabular}{l|c}
        objectives & acc.(\%) \\
        \hline
        VTC (Step 1) & 56.7 \\
        VTC+VTM (Step 2) & 60.3 \\
        VTC+VTM+MLM & 66.5 \\
        VTC+VTM+MLM+MVM & \textbf{67.5}
    \end{tabular}
\end{table}

\textit{Takeaway \#3: For the remaining experiments, we use VTC, VTM, MLM as our pretraining objectives.}

\subsection*{Step 4: Pretraining Data}

In this section, we analyze the effect of (i) the pretraining data, and (ii) pretraining protocols.

\xhdr{Datasets.} Several recent methods~\cite{bain2021frozen,fu2021violet} suggest that jointly pretraining on image and video data can lead to better performance. To investigate this, we consider an additional image-based CC3M~\cite{sharma2018conceptual} consisting of 3M image-text pairs. Specifically, we experiment with pretraining our framework on the (i) image-only (CC3M), (ii) video-only (WebVid2M), and (iii) joint image and video (CC3M + WebVid2M) datasets. When pretraining on images, we replace our previously introduced temporal attention module with an identity connection. This enables our model to be easily applied to both images and videos.

As shown in the Table below and Fig.~\ref{fig:expansion}, training on videos is more beneficial than training on images (\textbf{+2.7\%}), which makes sense as all of our downstream applications involve video. Furthermore, we observe that jointly pretraining on both images and videos leads to an additional \textbf{2.7\%} boost in performance. This suggests that the spatial and temporal cues are complementary and that a stronger spatial representation can boost VidL performance.
\begin{table}[h]
    \centering
    \small
    \begin{tabular}{l|cccc}
         & Images & Videos & Images+Videos \\
        \hline
        acc.(\%) & 64.8 & 67.5 & \textbf{70.2} \\
    \end{tabular}
\end{table}

\xhdr{The Number of Input Frames for Pretraining.} 
Prior approaches~\cite{bain2021frozen,fu2021violet,zellers2021merlot,wang2022omnivl} use a different number of input frames for pretraining (i.e., from 1 to 16). Thus, we next study how many frames are needed for effective VidL pretraining. 
The models are pretrained jointly on image and video (CC3M + WebVid2M) datasets.
From the Table below and Fig.~\ref{fig:expansion}, we observe that multi-frame pretraining using 4 frames leads to \textbf{1.7\%} improvement compared to a single-frame pretraining. However, we also observe that the performance saturates with 4-frame inputs while the computational cost of pretraining with more frames increases significantly. In particular, we note that pre-training with 4-frame inputs leads to a speedup of \textbf{2.5$\times$} compared to pretraining with 16-frame inputs. Thus, our finding is useful as it can save lots of computing power and speed up the development of future research.

\begin{table}[h]
    \centering
    \small
    \begin{tabular}{l|cccc}
         & 1 frame & 4 frames & 8 frames & 16 frames \\
        \hline
        acc.(\%) & 68.5 & \textbf{70.2} & \textbf{70.2} & \textbf{70.2} \\
        speedup & \textbf{4.6}$\times$ & $2.5\times$ & $1.7\times$ & $1\times$ \\
    \end{tabular}
\end{table}

\xhdr{Multi-stage Curriculum Pretraining.} Lastly, we also validate the necessity of multi-stage curriculum pretraining, which was used in several prior VidL approaches~\cite{bain2021frozen,wang2022omnivl}.
Specifically, we experiment with two different pretraining protocols: (i) a two-stage pretraining that first trains a model for 10 epochs using single frames, and then uses a multi-frame training for 5 additional epochs using 4-frame inputs, and (ii) a three-stage pretraining that builds on (i) by adding a third stage where the model is trained for additional 3 epochs using 8-frame inputs. 
The model is pretrained jointly on image and video datasets.
Our results in the Table below and Figure~\ref{fig:expansion}, indicate that multi-stage pretraining does not lead to any significant boost in performance, which is contrary to the findings of prior approaches~\cite{bain2021frozen,wang2022omnivl}. We conjecture that this might happen because prior approaches~\cite{bain2021frozen,wang2022omnivl} train their model for only several epochs at each stage whereas we train it until convergence (10 epochs for the first stage).
We also note that compared to the 4-frame one-stage pretraining (described above), the two-stage $1\to4$ has a comparable pretraining cost as the latter model is trained for more epochs.
\begin{table}[h]
    \centering
    \small
    \begin{tabular}{l|ccccc}
        frames & $4$ & $1\to4$ & $1\to 4 \to 8$ & $4\to 8$ \\
        \hline
        acc.(\%) & 70.2 & 69.6 & 69.5 & \textbf{70.4} \\
        speedup & \textbf{1.7}$\times$ & \textbf{1.7}$\times$ & 1.2$\times$ & 1$\times$
    \end{tabular}
\end{table}

\textit{Takeaway \#4: We adopt a single-stage pretraining on joint image and video datasets while using 4-frame inputs.}

\subsection*{Step 5: Finetuning \& Inference} 

Existing methods typically use the same number of frames either between pretraining and finetuning~\cite{bain2021frozen,zellers2021merlot,lei2022revealing} or between finetuning and inference~\cite{zellers2021merlot,fu2021violet,wang2022omnivl}. Here, we study whether we can use a different number of frames at different phases.

\xhdr{Finetuning.} We experiment with finetuning our 4-frame pretrained model with ${K=1, 4, 8, 12, 24, 32}$-frame inputs while using $M$ frames during inference. We use $M=12$ for all $K \le 12$ and $M=K$ for $K > 12$ as we found inference with more frames leads to higher performance. Based on the results in the Table below, we observe that while finetuning with more frames leads to higher accuracy (\textbf{70.5}\%) the performance saturates with about 12 frames. We also note that finetuning with a single-frame input is \textbf{22.4}$\times$ faster than with 32-frame inputs but has a \textbf{5\%} lower accuracy. On the other hand, finetuning with 12-frame inputs yields only \textbf{0.3\%} lower accuracy but \textbf{2.6}$\times$ speedup compared to finetuning with 32-frame inputs. Therefore, due to the favorable accuracy-cost tradeoff, we finetune most of our models with 12-frame inputs.

\begin{table}[h]
    \centering
    \small
    \begin{tabular}{l|cccccc}
        \# frames & 1 & 4 & 8 & 12 & 24 & 32 \\
        \hline
        acc.(\%) & 65.5 & 68.1 & 69.2 & 70.2 & 70.1 & \textbf{70.5} \\
        speedup & \textbf{22.4}$\times$ & 7.1$\times$ & 3.9$\times$ & 2.6$\times$ & 1.5$\times$ & 1.0$\times$
    \end{tabular}
\end{table}

\xhdr{Inference.} Next, we also experiment with using {12, 24, 32, 64} frames for testing our 4-frame pretrained and 12-frame finetuned model.
In the table below, we report the averaged accuracies on the DiDeMo (D) / ActivityNet (A) datasets, which contain longer videos. Using more frames for inference is beneficial, but the performance also saturates quickly, and the inference speed slows down rapidly.

\begin{table}[h]
    \centering
    \small
    \begin{tabular}{l|cccccc}
        \# frames & 12 & 24 & 32 & 64 \\
        \hline
        D/A acc.(\%) & 73.4/70.4 & 73.0/72.1 & 72.7/72.6 & \textbf{73.8}/\textbf{72.8} \\
        speedup & \textbf{10.6}$\times$ & 3.1$\times$ & 2.1$\times$ & 1$\times$ 
    \end{tabular}
\end{table}

\textit{Takeaway \#5: Considering the trade-off between computational cost and accuracy, we use 12 frames for finetuning and inference on all datasets except ActivityNet. On ActivityNet, we use 12 and 32 frames for finetuning and inference.}

\subsection*{Step 6: Scaling Up}

As our last step, we investigate scaling up the pretraining data and the model size. 

\xhdr{Pretraining Data.} For the pre-training data, we experiment with (a)  adding 12M images from CC12M for a \textbf{17M Corpus}, and (b) additional 10M videos from WebVid10M for a \textbf{25M Corpus}. The results in the Table below and in Figure~\ref{fig:expansion} indicate that scaling our corpus from $5M \rightarrow 17M$ improves the downstream VidL performance by \textbf{2.2\%}. Furthermore, scaling the corpus from $17M \rightarrow 25M$ leads to an additional boost of \textbf{1.2\%}. 
\begin{table}[h]
    \centering
    \small
    \begin{tabular}{l|ccc}
        \# corpus & 5M & 17M & 25M \\
        \hline
        acc.(\%) & 70.2 & 72.4 & \textbf{73.6} 
    \end{tabular}
\end{table}

\xhdr{Model Size.} In the Table below, we also experiment with scaling the video encoder ($\text{ViT}_{base}\to \text{ViT}_{large}$) or text encoder ($\text{BERT}_{base}\to \text{BERT}_{large}$).
Due to the large computational cost, we could only conduct these experiments on the 5M corpus.
We report that scaling the vision encoder brings larger improvement ( \textbf{+3.0\%}) than scaling the text encoder (\textbf{+1.0\%}).
\begin{table}[h]
    \centering
    \small
    \begin{tabular}{l|ccc}
        encoders & base & $\text{ViT}_{large}$ & $\text{BERT}_{large}$ \\
        \hline
        acc.(\%) & 70.2 & \textbf{73.2} & 71.2 
    \end{tabular}
\end{table}

\textit{Final Takeaway: Our final scaled-up \Modelname~model improves the initial image transformer baseline by \textbf{23.2\%}.}

\subsection*{Other Useful Empirical Tips} 

\begin{table}
    \centering
    \small
    \begin{tabular}{lcccc}
    \toprule
       Visual Encoder  & MSR-VTT & DiDeMo & ANet & Avg. \\
       \midrule
        ViT\cite{dosovitskiy2020image,bao2021beit} & \textbf{64.5} & \textbf{75.0} & 72.9 & \textbf{70.8} \\
        VideoSwin\cite{liu2022video} & 61.1 & 73.1 & \textbf{73.4}& 69.2 \\
    \bottomrule
    \end{tabular}
    \caption{We study the performance of Isotropic (ViT) vs. Pyramid (VideoSwin) vision encoders. Based on these results, we observe that ViT outperforms VideoSwin by 1.6\% on an averaged R@\{1,5,10\} on MSR-VTT, DiDeMo and ActivityNet-Captions. We experiment with 4 frames using our final model on the 5M corpus.}
    \label{tab:tips_enc}
\end{table}

\xhdr{Isotropic vs Pyramid-based Vision Encoder.} 
Pyramid-style ViTs that use downsampling along the spatial dimension (e.g., Swin\cite{liu2021swin}, MViT\cite{fan2021multiscale}) have shown stronger performance than isotropic ViTs (vanilla ViT) on many image/video classification tasks. Thus, several recent VidL approaches~\cite{fu2021violet,li2022lavender,fu2022empirical} adopt pyramid ViTs as their vision encoders.
However, in our study, we find that isotropic ViTs tend to have better performance.
Specifically, in Tab.~\ref{tab:tips_enc}, we show that a ViT-based encoder outperforms VideoSwin by \textbf{1.6\%}. We hypothesize that this might happen because isotropic ViTs preserve more fine-grained spatial information needed for various VidL tasks.

\xhdr{A Linear Scaling Rule.}  Linear scaling strategy~\cite{goyal2017accurate} has been extensively used for large-scale pretraining on image/video classification tasks. However, in our setting, we observed that the linear scaling rule leads to similar or worse results  (See Table~\ref{tab:tips_lr}). Therefore, for all of our experiments, we use a fixed learning rate (1e-4) for all batch sizes.

\xhdr{Initialization.} We also found that the initialization of various modules in our model is critical for good VidL performance. In particular, we note that to make MLM and MVM pretraining objectives effective, we need to use text and video encoders pretrained with these objectives in a self-supervised manner (e.g., BERT~\cite{devlin2018bert} and BEIT~\cite{bao2021beit} respectively). Otherwise, the performance will drop significantly ($\sim$\textbf{5}\% averaged R@1,5,10 accuracy drop on MSR-VTT, DiDeMo, ActivityNet datasets).

\begin{table}
    \centering
    \small
    \begin{tabular}{lcccccc}
    \toprule
    Batch Size & 512 & 1024 & 1024 & 2048 & 2048 & 2048 \\
    \midrule
    LR ($\times$1e-4) & 1 & 1 & 2 & 1 & 2 & 4 \\
    \midrule
    Accuracy & 68.2 & 68.2 & 68.2 & \textbf{68.5} & 68.3 & 67.4 \\
    \bottomrule
    \end{tabular}
    \caption{We investigate the effectiveness of a scaled learning rate rule~\cite{goyal2017accurate} using averaged downstream accuracy on MSR-VTT, DiDeMo, and ActivityNet-Captions. The learning rate 1e-4 works best for various batch sizes. We experiment with 1-frame inputs using our final model on the the 5M corpus.
    }
    \label{tab:tips_lr}
\end{table}

\begin{table*}
    \centering
    \small
    \setlength{\tabcolsep}{4pt}
    \begin{tabu}{lrccccccccccccccc}
        \toprule
        \multirow{2}{*}{\bf Method} & \multicolumn{3}{c}{\bf Pretrain} & \multicolumn{4}{c}{\bf MSRVTT} & \multicolumn{4}{c}{\bf DiDeMo} & \multicolumn{4}{c}{\bf ActivityNet-Captions} & \multirow{2}{*}{\bf Avg}\\
        \cmidrule{2-4}\cmidrule(lr){5-8} \cmidrule(lr){9-12} \cmidrule(lr){13-16}
        & \#Data & \#Frames & Time & R1 & R5 & R10  & Avg & R1 & R5 & R10  & Avg & R1 & R5 & R10 & Avg \\
        \midrule

        ClipBERT~\cite{lei2021less}  & 5.4M & 1 & 32 & 22.0 & 46.8 & 59.9 & 42.9 & 20.4 & 48.0 & 60.8 & 43.1 & 21.3 & 49.0 & 63.5 & 44.6 & 43.5 \\
        VideoCLIP~\cite{xu2021videoclip}  & 136M & 960 & 8 & 30.9 & 55.4 & 66.8 & 51.0 & - & - & - & - & - & - & - & - & -\\
        Frozen~\cite{bain2021frozen}  & 5M & $1\to4$ & $35^*$ & 31.0 & 59.5 & 70.5 & 53.7 & 34.6 & 65.0 & 74.7 & 58.1 & - & - & - & - & - \\
        ALPRO~\cite{li2022align}  & 5M & 8 & $24^*$ & 33.9 & 60.7 & 73.2 & 55.9 & 35.9 & 67.5 & 78.8 & 60.7 & - & - & - & - & -\\
        VIOLET~\cite{fu2021violet}  & 138M & 4 & 83 & 34.5 & 63.0 & 73.4 & 57.0 & 32.6 & 62.8 & 74.7 & 56.7 & - & - & - & - & -\\
        All-in-one~\cite{wang2022all} & 138M & 3 & 448 & 37.9 & 68.1 & 77.1 & 61.0 & 32.7 & 61.4 & 73.5 &55.9 & 22.4 & 53.7 & 67.7 & 47.9 & 54.9 \\
        LAVENDER~\cite{li2022lavender} & 30M & 4 & 640 & 40.7 & 66.9 & 77.6 & 61.7 & \underline{53.4} & 78.6 & 85.3 & 72.4 & - & - & - & - & - \\
        Singularity~\cite{lei2022revealing} & 17M & $1\to4$ & 29 & 42.7 & 69.5 & 78.1 & 63.4 & 53.1 & \underline{79.9} & \underline{88.1} & \underline{73.7} & \underline{48.9} & \underline{77.0} & \underline{86.3} & \underline{70.7} & \underline{69.3}\\
        OmniVL~\cite{wang2022omnivl} & 17M & $1\to8$ & $169^*$ & 47.8 & \textbf{74.2} & \textbf{83.8} & \textbf{68.6} & 52.4 & 79.5 & 85.4 & 72.4 & - & - & - & - & - \\ 
        \rowfont{\color{Gray}}
        CLIP4Clip~\cite{luo2022clip4clip} & 400M & 1 & $768^*$ & 44.5 & 71.4 & 81.6 & 65.8 & 42.8 & 68.5 & 79.2 & 63.5 & 40.5 & 72.4 & 83.4 & 65.4 & 64.9\\
        \rowfont{\color{Gray}}
        ECLIPSE~\cite{lin2022eclipse} & 400M & 1 & $768^*$ & - & - & - & -- & 44.2 & - & - & - & 45.3 & 75.7 & 86.2 & 69.1 & -\\
        \rowfont{\color{Gray}}
        CLIP-Hhiker~\cite{bain2022clip} & 400M & 1 & $768^*$ & 47.7 & 74.1 & 82.9 & 68.6 & - & - & - & - & 44.0 & 74.9 & 86.1 & 68.3 & - \\
        \rowfont{\color{Gray}}
        CLIP-ViP~\cite{xue2022clip} & 500M & $1\to12$ & $984^*$ & 54.2 & 77.2 & 84.8 & 72.1 & 50.5 & 78.4 & 87.1 & 72.0 & 53.4 & 81.4 & 90.0 & 74.9 & 73.0\\
        \midrule
        \multirow{3}{*}{\Modelname} & 5M & \multirow{3}{*}{4} & 15 & 43.8 & 70.3 & 79.5 & 64.5 & 54.6 & 81.3 & 89.0 & 75.0 & 51.1 & 79.2 & 88.4 & 72.9 & 70.8 \\
        & 17M & & 38 & 45.3 & 69.9 & 79.6 & 64.9 & 59.2 & 84.1 & 89.5 & 77.6 & 54.4 & 80.7 & 89.0 & 74.7 & 72.4\\
        & 25M & & 82 & 46.5 & 71.5 & 80.4 & 66.1 & \textbf{61.2} & 85.8 & 91.0 & \textbf{79.3} & 55.0 & 81.4 & 89.7 & 75.4 & 73.6\\
        \midrule
        \Modelname-L & 25M & 4 & 178 & \textbf{48.8} & \underline{72.4} & \underline{82.2} & \underline{67.8} & 59.8 & \textbf{86.6} & \textbf{91.5} & \textbf{79.3} & \textbf{55.9} & \textbf{82.3} & \textbf{90.9} & \textbf{76.4} & \textbf{74.5}\\
        \bottomrule
    \end{tabu}
    \caption{Comparison to the state-of-the-art text-to-video retrieval methods on MSRVTT, DiDeMo and AcitivityNet-Captions. Pretraining time is measured in V100 GPU days, where * means our estimated time based on FLOPs, pretraining data, and the number of epochs for the methods that do not report their pretraining time. \Modelname~uses ViT-B/16 while \Modelname-L uses ViT-L/16 as video encoders. For fair comparisons, we de-emphasize the CLIP-based methods since they use a lot more pretraining data than all the other approaches. Our results indicate that \Modelname~achieves competitive or even better than state-of-the-art results on all three datasets while also being simple and efficient.
    }
    \label{tab:retrieval}
\end{table*}

\begin{table}[htpb]
    \centering
    \small
    \setlength{\tabcolsep}{3pt}
    \begin{tabu}{lrccccc}
        \toprule
        \multirow{2}{*}{\bf Method} & \multirow{2}{*}{\bf \#PT} & \multicolumn{2}{c}{\bf SSv2-label} & \multicolumn{2}{c}{\bf SSv2-template} & \multirow{2}{*}{\bf Avg} \\
        \cmidrule(lr){3-4} \cmidrule(lr){5-6}
        & & R1 & R5 & R1 & R5 & \\
        \midrule
        \rowfont{\color{Gray}}
        CLIP4Clip~\cite{luo2022clip4clip} & 400M & 43.1 & 71.4 & 77.0 & 96.6 & 77.9 \\
        Singularity~\cite{lei2022revealing} & 17M & 47.4 & 75.9 & 77.6 & 96.0 & 80.0 \\
        \midrule
        \multirow{3}{*}{\Modelname} & 5M & 51.2 & 78.8 & 82.2 & 98.9 & 82.7 \\
         & 17M & 53.0 & 80.8 & \textbf{86.2} & 99.4 & \textbf{84.6} \\
         & 25M & \textbf{53.1} & \textbf{81.8} & 83.3 & \textbf{100} & 84.4 \\
        \bottomrule
        
    \end{tabu}
    \caption{Comparison with state-of-the-art text-to-video retrieval methods on the temporally-heavy SSv2-Label~\cite{lei2022revealing} and SSv2-Template datasets~\cite{lei2022revealing}. \#PT denotes the amount of pretraining data. 
    Averaged numbers are the average of Recal@\{1,5,10\} on these two datasets.
    CLIP-based models are de-emphasized for fairer comparisons. Based on these results, we observe that \Modelname~achieves the best performance, which demonstrates its ability to reason about complex temporal dependencies in the video data.
    }
    \label{tab:retrieval_ssv2}
\end{table}

\begin{table}
    \centering
    \small
    \setlength{\tabcolsep}{2pt}
    \begin{tabu}{lrcccc}
        \toprule
        \bf Method & \bf \#PT & \bf ANet & \bf MSR-QA & \bf MSR-MC & \bf TVQA \\
        \midrule
        ClipBERT~\cite{lei2021less} & 0.2M & - & 37.4 & 88.2 & - \\
        ALPRO~\cite{li2022align} & 5M & - & 42.1 & - & - \\
        JustAsk~\cite{yang2021just} & 69M & 38.9 & 41.5 & - & - \\
        VideoCLIP~\cite{xu2021videoclip} & 136M & - & - & 92.1 & - \\
        All-in-one~\cite{wang2022all} & 138M & - & 44.3 & 92.0 & - \\
        MERLOT~\cite{zellers2021merlot} & 180M & 41.4 & 43.1 & 90.9 & 78.7 \\
        VIOLET~\cite{fu2021violet} & 138M & - & 43.9 & 91.9 & - \\
        Singularity~\cite{lei2022revealing} & 17M & 44.1 & 43.9 & 93.7 & - \\
        OmniVL~\cite{wang2022omnivl} & 17M & - & 44.1 & - & - \\
        HERO~\cite{li2020hero} & 7.5M & - & - & - & 74.2 \\
        \rowfont{\color{Gray}}
        FrozenBiLM~\cite{yang2022zero} & 400M & 43.2 & 47.0 & - & 82.0 \\
        \midrule
        \multirow{3}{*}{\Modelname} & 5M & 44.2 & 43.6 & 95.4 & \textbf{79.0} \\
        & 17M & 44.6 & 43.8 & 93.8 & 78.8 \\
        & 25M & \textbf{44.7} & \textbf{44.6} & \textbf{95.5} & \textbf{79.0} \\
        \bottomrule
        
    \end{tabu}
    \caption{Comparison with state-of-the-art video question-answering methods on ActivityNet-QA (ANet), MSRVTT-QA (MSR-QA), MSRVTT-MC (MSR-MC) and TVQA. \#PT denotes the amount of pretraining data. We gray out FrozenBiLM~\cite{yang2022zero} as it is much larger than our model (1.2B vs 207M parameters). We observe that \Modelname~achieves competitive results across all four of these datasets.  
    }
    \label{tab:vqa}
\end{table}

\begin{table}
    \centering
    \small
    \begin{tabular}{lccc}
        \toprule
        Method & TimeSformer~\cite{gberta_2021_ICML} & OmniVL~\cite{wang2022omnivl} & \Modelname \\
        \midrule
        Top-1 acc. & 78.0 & 79.1 & \textbf{80.1} \\
        \bottomrule
    \end{tabular}
    \caption{Results on Kinetics-400~\cite{kay2017kinetics} for action recognition task. All models use the same TimeSformer architecture~\cite{gberta_2021_ICML}. Our \Modelname~approach outperforms both the TimeSformer~\cite{gberta_2021_ICML} and OmniVL~\cite{wang2022omnivl} baselines by $2.1\%$ and $1.0\%$ respectively. These results indicate the benefits of our VidL pretraining recipe.}
    \label{tab:act_rec}
\end{table}

\section{Experimental Results} 
We validate our \Modelname~ recipe on two mainstream VidL tasks. See implementation details in Appx.~\ref{appendix:implementation} and dataset descriptions in Appx.~\ref{appendix:datasets}. 

\xhdr{Text-to-Video Retrieval.} We compare our results with existing methods on three spatially-biased datasets MSR-VTT, DiDeMo, and ActivityNet and two temporally-heavy datasets, SSv2-label, and SSv2-template as shown in Tab.~\ref{tab:retrieval} and Tab.~\ref{tab:retrieval_ssv2} respectively.
Our method outperforms previous methods by a large margin on multiple datasets, achieving averaged accuracies of 79.3\% (\textbf{+5.6\%}), 75.4\% (\textbf{+4.7\%}), 84.6\% (\textbf{+4.6\%}) on DiDeMo, ActivityNet-Captions and SSv2 respectively.
Our results on MSR-VTT are worse (\textbf{66.5\%} vs. \textbf{68.6\%}) than OmniVL~\cite{wang2022omnivl} but our pretraining framework is significantly cheaper (i.e., \textbf{82} vs. \textbf{169} V100 GPU days). We also note that our method is significantly cheaper than other top-performing approaches including LAVENDER~\cite{li2022lavender}, All-in-one~\cite{wang2022all}, and CLIP-ViP~\cite{xue2022clip} (\textbf{82} vs. \textbf{640, 448, 984} V100 GPU days for pretraining respectively).  Additionally, our cheapest \Modelname~variant requires only $15$ V100 GPU days for pre-training, which is the second cheapest model among all listed approaches, and it still achieves competitive results on all three benchmarks. Furthermore, compared to the other leading VidL approaches such as OmniVL and Singularity, which rely on a multi-stage curriculum pretraining, our framework is simpler since it can be trained in a single stage. Lastly, our results on the SSv2 dataset in Table~\ref{tab:retrieval_ssv2} indicate that \Modelname~performs very well not only on spatially-biased datasets but also on temporally-heavy datasets, which require sophisticated temporal modeling capabilities.  For fairer comparisons, we de-emphasize CLIP-based methods since they use a lot more pre-training data. 

\xhdr{Video Question-Answering.}  In Table~\ref{tab:vqa}, we also present our results for the video question-answering task on ActivityNet-QA~\cite{yu2019activitynet}, MSRVTT-QA~\cite{xu2017video}, MSRVTT-MC~\cite{yu2018joint} and TVQA~\cite{lei2018tvqa}. Our results indicate that compared to prior state-of-the-art approaches, \Modelname~achieves competitive results across all four of these datasets. In particular, our method outperforms existing approaches by \textbf{0.6\%} on ActivityNet-QA, \textbf{0.3\%} on MSRVTT-QA, \textbf{3.4\%} on MSRVTT-MC and \textbf{0.3\%} on TVQA. For fair comparison, we de-emphasize FrozenBiLM~\cite{yang2022zero}, since it is a lot larger than our model (1.2B vs. 201M parameters) and uses a lot more pretraining data (400M vs. 25M).

\xhdr{Action Recognition.} We finetune our pretrained video encoder on Kinetics-400~\cite{kay2017kinetics} directly using TimeSformer~\cite{gberta_2021_ICML} codebase with exactly the same hyperparameters as in~\cite{gberta_2021_ICML}. As shown in Table~\ref{tab:act_rec}, our video encoder outperforms TimeSformer~\cite{gberta_2021_ICML} and OmniVL~\cite{wang2022omnivl} by \textbf{2.1}\% and \textbf{1.0}\% respectively with all models using exactly the same architecture~\cite{gberta_2021_ICML}. This indicates the usefulness of our VidL pretraining recipe for a pure video understanding task.

\section{Conclusion}
In this work, we demystify  the importance of various components used in modern VidL framework design. Throughout our empirical study, we find that temporal modeling, multimodal fusion, masked modeling pretraining objectives, and joint training on images and videos are critical for good performance on the downstream VidL understanding tasks. Our empirical insights enable us to develop a step-by-step recipe for effective video-language (VidL) pretraining, which leads to a highly performant VidL model, dubbed \Modelname. Compared to the existing VidL approaches, our method achieves competitive or even better results on 9 VidL benchmarks while also being simpler and more efficient. While our paper does not provide any novel individual contributions, we believe that our empirical insights and our VidL pretraining recipe will be useful and help advance further research in the VidL domain.

\paragraph{Acknowledgements.} We thank Yan-Bo Lin,  Md Mohaiminul Islam, Avinash Madasu and Maitrey Gramopadhye for helpful discussions. This work was supported by the Sony Faculty Innovation award, Lilly Endowment, Inc. via Indiana University Pervasive Technology Institute, Laboratory for Analytic Sciences via NC State University and NSF-AI Engage Institute DRL211263.

\section*{Appendix}
\appendix

\section{Implementation Details}
\label{appendix:implementation}

\xhdr{Positional Embeddings.} We use learnable absolute temporal positional embeddings as in~\cite{bain2021frozen} and relative spatial positional embeddings as in~\cite{bao2021beit}. 
The temporal positional embeddings are applied after patchifying the tokens, while the relative spatial positional embeddings are applied at each Transformer layer.
When adapting the pretrained model to downstream tasks with more frames, we use zero-padding for the temporal positional embeddings as in~\cite{bain2021frozen}. When adapting to higher spatial resolutions, we linearly interpolate the spatial positional embeddings.

\xhdr{Video Retrieval.} We finetune the pretrained model with VTC and VTM losses. During inference, we follow~\cite{li2022blip,li2021align} to first select top-$K$ ($K=128$ in our experiments) candidates based on the video-text similarity scores of the unimodal encoders and then re-rank these candidates by calculating their pairwise VTM scores.

\begin{figure}[b]
    \centering
    \includegraphics[width=0.8\linewidth]{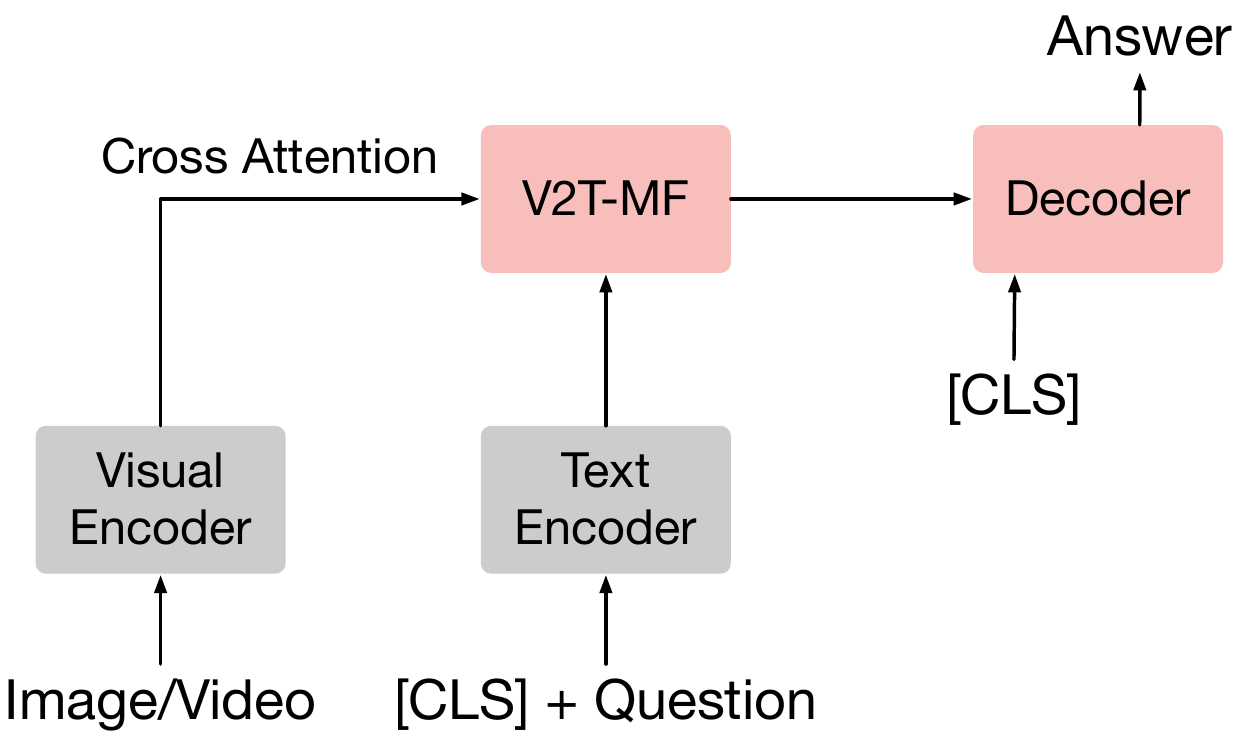}
    \caption{Our architecture for the open-ended question-answering task. The decoder uses the same architecture as our video-to-text multimodal fusion (V2T-MF) module and is initialized with the pretrained V2T-MF's weights.}
    \label{fig:vqa}
\end{figure}

\xhdr{Open-ended Question-Answer.} Following~\cite{lei2022revealing,wang2022omnivl,li2021align}, we formulate this task as a text generation task. As shown in Fig.~\ref{fig:vqa}, we add a decoder that takes the multimodal encoder's outputs as the cross attention key and value to generate the answers. The decoder starts with a [CLS] token and ends when a [SEP] token is generated. The decoder has the same architecture as the multimodal encoder and is initialized with the pretrained multimodal encoder's weights.
The model is optimized using the averaged cross-entropy loss of each token between the generated answer and the ground truth answer.
For a fair comparison with prior works~\cite{lei2022revealing, wang2022all, wang2022omnivl}, we constrain the decoder to generate from the 3128 most common answers~\cite{lei2022revealing} during inference. 

\begin{table*}
    \centering
    \small
    \begin{tabular}{lcccccccc}
        \toprule
        \multirow{2}{*}{Config} & \multirow{2}{*}{Pretraining} & \multicolumn{4}{c}{Video Retrieval} & \multicolumn{3}{c}{Video QA} \\
        \cmidrule(lr){3-6} \cmidrule(lr){7-9}
        & & MSRVTT & DiDeMo & ANet & SSv2-* & ANet & MSRVTT-QA & TVQA \\
        \midrule
        optimizer & \multicolumn{8}{c}{AdamW~\cite{loshchilov2017decoupled}} \\
        optimizer options & \multicolumn{8}{c}{$\beta_1=0.9, \beta_2=0.999$} \\
        weight decay & \multicolumn{8}{c}{0.02}\\
        learning rate schedule & \multicolumn{8}{c}{cosine decay~\cite{loshchilov2016sgdr}}\\
        init learning rate & 1e-4 & 1e-5 & 1e-5 & 1e-5 & 1e-4 & 1e-5 & 1e-5 & 1e-5 \\
        min learning rate & 1e-6 & 1e-6 & 1e-6 & 1e-6 & 1e-5 & 1e-6 & 1e-6 & 1e-6 \\
        \midrule
        spatial resolution & \multicolumn{8}{c}{$224\times 224$} \\
        augementation & \multicolumn{8}{c}{random resize, crop, horizontal flip} \\
        \midrule
        \# epochs & 10 & 5 & 10 & 10 & 10 & 10 & 10 & 10 \\
        \# warmup epochs & 1 & 0.5 & 0.5 & 0.5 & 0 & 0 & 0 & 0 \\
        batch size $\times$ \# GPUs & $64\times \{8,32\}$ & $32\times 4$ & $32\times 1$ & $32\times 1$ & $32\times 2$ & $32\times 1$ & $32\times 1$ & $32\times 1$\\
        \midrule 
        \# training frames & 4 & 12 & 12 & 12 & 12 & 12 & 12 & 12 \\
        \# inference frames & 4 & 12 & 12 & 32 & 12 & 32 & 12 & 12 \\
        \bottomrule
        
    \end{tabular}
    \caption{Hyper-parameters for pretraining, and downstream tasks. SSv2-* means SSv2-Template and SSv2-Label datasets. We pretrain on 8 GPUs for C2M and C5M, 32 GPUs for C17M and C25M.
    }
    \label{tab:hypers}
\end{table*}

\xhdr{Multiple-Choice Question-Answering.} For Multiple-Choice QA, we follow~\cite{lei2022revealing,wang2022omnivl,li2021align} and convert it to the text-to-video retrieval task. Specifically, for each question and $m$ candidate answers, we generate $m$ sentences by concatenating the question with each candidate's answer. We then rank these sentences by ensembling the retrieval model's video-text similarity and pairwise VTM scores. The ensembling weights are set to 0.3 for the similarity score and 0.7 for the VTM score.

\xhdr{Inference with More Frames.} Following~\cite{lei2022revealing}, we perform inference using more frames than our finetuned model. Specifically, we first linearly interpolate the temporal positional embeddings in the video encoder. Then all the visual tokens are concatenated and fed to the multimodal encoder.

 \xhdr{Pretraining Datasets.}
 \label{sec:method:impl}
 As discussed in the main draft, in Steps 1-3 of our recipe, we pretrain our model on a 2M WebVid-2M~\cite{bain2021frozen} corpus.
 For Steps 4-5, we use a joint image-video corpus consisting of 3M images from CC3M~\cite{sharma2018conceptual} and 2M videos from WebVid-2M~\cite{bain2021frozen}.
 Lastly, in Step 6, we scale our pretraining data from $5M \rightarrow 17M \rightarrow 25M$.

 \xhdr{Model Details.} Our final \Modelname~uses a vision encoder based on ViT~\cite{dosovitskiy2020image} architecture initialized with $\text{BEIT}_{base}$~\cite{bao2021beit} weights, pretrained on ImageNet-21k. The additional temporal attention modules are randomly initialized and added before spatial attention in each Transformer block as in~\cite{gberta_2021_ICML}.
 As our text encoder, we use the first 9 layers of $\text{BERT}_{base}$~\cite{devlin2018bert}. The multimodal fusion encoder is our previously described V2T-MF module built using the last 3 layers of the same $\text{BERT}_{base}$ model. Our final pretraining objective is the sum of VTC, VTM and MLM losses.
 The hyperparameters are shown in Table~\ref{tab:hypers}.
 When doing multi-stage pretraining in Step 4 in the main draft, we set the initial learning rate of 5e-5 for stage 2 and 1e-6 for stage 3.
 Our model is implemented using PyTorch~\cite{paszke2019pytorch} with Mixed Precision Training~\cite{micikevicius2017mixed} and Gradient Checkpointing~\cite{chen2016training}.

 \xhdr{Training Time.} We train 2M and 5M corpus on $8\times$ RTX A5000 GPUs, which takes about 1 day and 1.8 days, respectively.
 For 17M and 25M, we train our model using $32\times$ A5000 GPUs, which takes 1.3 days and 3 days, respectively.
 For downstream tasks, the finetuning time ranges from 2-40 hours depending on the dataset size.
The speed of A5000 is $0.99\times$ as V100 and $0.5\times$ as the A100 according to Lambda's benchmark\footnote{\url{https://lambdalabs.com/gpu-benchmarks} fp16, bert\_base\_squad}.

\section{Dataset Descriptions} 
\label{appendix:datasets}

\xhdr{Pretraining.} We pretrain our model on three corpora: C5M, C17M and C25M, which we describe below.
{
\begin{itemize}
    \item \textbf{C5M (5M)}: WebVid-2M~\cite{bain2021frozen}, and CC3M~\cite{sharma2018conceptual}. It contains a total of 5.44M image/video and text pairs.
    \item \textbf{C17M (17M)}: C5M, COCO~\cite{chen2015microsoft}, Visual Genome~\cite{krishna2017visual}, SBU Captions~\cite{ordonez2011im2text}, and CC12M~\cite{changpinyo2021conceptual}. It contains a total of 18.41M image/video and text pairs.
    \item \textbf{C25M (25M)}: C17M, and WebVid-10M~\cite{bain2021frozen} (excluding 2M videos from WebVid-2M as WebVid-10M is a superset of WebVid-2M). It contains a total of 25.91M image/video and text pairs.
\end{itemize}
}

\xhdr{Text-to-Video Retrieval.} We evaluate our model on 3 spatially biased datasets MSR-VTT~\cite{xu2016msr}, DiDeMo~\cite{anne2017localizing}, ActivityNet- Captions~\cite{krishna2017dense} and 2 temporally-heavy datasets SSv2-Template~\cite{lei2022revealing}, SSv2-Label~\cite{lei2022revealing}. 
{
\begin{itemize}

    \item \textbf{MSRVTT~\cite{xu2016msr}} contains 10K YouTube videos with duration between 10-30 seconds and 200k captions. Following ~\cite{yu2018joint,bain2021frozen}, we train on 9K videos and report results on 1K-A test set.
    \item \textbf{DiDeMo~\cite{anne2017localizing}} contains 10K Flicker videos with 41K captions. Following~\cite{lei2021less,lei2022revealing,li2022lavender}, we only keep the first 30 seconds of each video and evaluate paragraph-to-video retrieval, where all the descriptions for a video are concatenated to form a single query.
    \item \textbf{ActivityNet-Captions~\cite{caba2015activitynet}} contains 20K YouTube videos with 100K captions. Following~\cite{luo2022clip4clip,lei2022revealing}, we train on the train set with 10K videos and evaluate on the val set with 4.9K videos and evaluate paragraph-to-video retrieval.
    \item \textbf{SSv2-Template~\cite{lei2022revealing}} contains 169K videos for training and 2K videos for evaluation from dataset SSv2~\cite{goyal2017something}.
    The queries are 174 template (e.g., ``Holding [something] next to [something]") in SSv2. In the 2K test set, each template has 12 videos.
    \item \textbf{SSv2-Label~\cite{lei2022revealing}} contains the same videos for train/test as in SSv2-Template except that the text queries are the annotated labels (e.g., ``holding potato next to vicks vaporub bottle") in SSv2.
\end{itemize}
}

\xhdr{Video Question Answering.} We evaluate on two open-ended QA datasets ActivityNet-QA, MSRVTT-QA and two multiple-choice QA dataset MSRVTT-MC, TVQA.
{
\begin{itemize}
    \item \textbf{ActivityNet-QA}~\cite{yu2019activitynet} contains 58K open-ended questions on 5.8K sampled videos from ActivityNet~\cite{krishna2017dense}.
    \item \textbf{MSRVTT-QA}~\cite{xu2017video} contains 244K open-ended questions on 10K MSRVTT videos.
    \item \textbf{MSRVTT-MC}~\cite{yu2018joint} contains 3K sampled videos with one multiple choice question for each video with 5 candidates. We evaluate the performance using the retrieval model finetuned on MSRVTT 7K training set.
    \item \textbf{TVQA}~\cite{lei2018tvqa} contains 22K video clips and 153K multiple-choice questions focused on popular TV shows. We use the official train/val/test splits and reports results on the test set.
\end{itemize}
}

\section{Additional Quantitative Results}
In this section, we present additional quantitative results on temporal modeling.

\begin{figure*}
    \centering
    \includegraphics[width=0.8\textwidth]{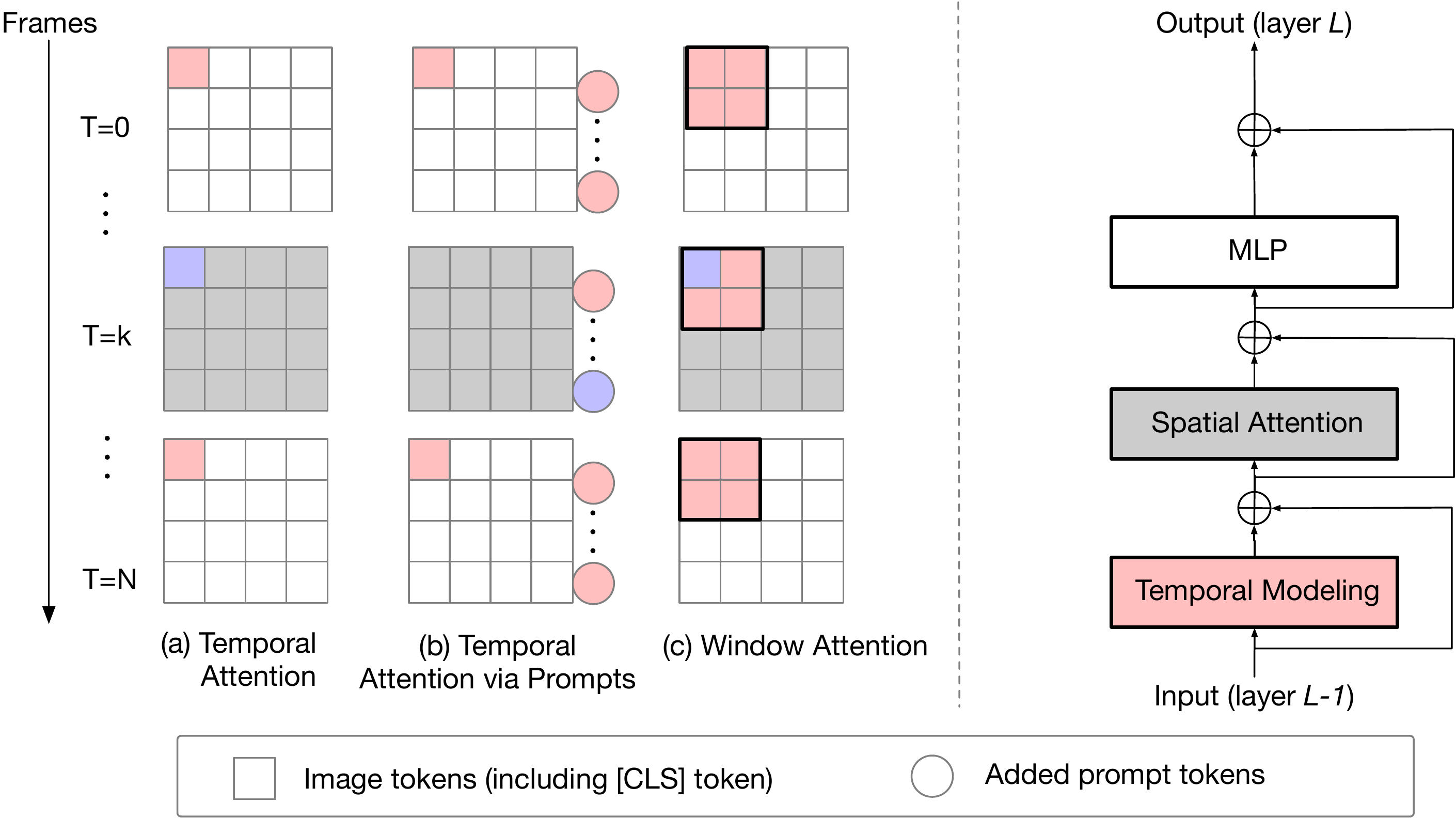}
    \caption{
    \textbf{Left}: Comparison of different attention mechanisms. The \textcolor{blue}{query} token will first attend to \textcolor{Salmon}{tokens in other frames} via temporal attention machnism and then attend to \textcolor{gray}{tokens in the same frames} via spatial attention.
    In Temporal Attention via Prompts, only spatial attention is applied to image tokens.
    \textbf{Right}: The temporal modeling blocks are inserted before the spatial attention in each ViT layer.
    }
    \label{fig:temporal_modeling}
\end{figure*}

\subsection{Additional Ablation Studies}

\label{exp:ablation:mlm}
\xhdr{MLM masking ratio.} We found a larger masking ratio (50\%) for the MLM objective is more helpful for VidL pretraining, compared to the 15\% masking ratio used in BERT~\cite{devlin2018bert}. We conjecture that we can use a higher mask ratio than text-only BERT because our model incorporates complementary video cues.

\xhdr{Analysis on More Tasks/Datasets.} In Tab.~\ref{tab:ablation_more_datasets}, we further evaluate our recipe on VidQA on MSRVTT-QA~\cite{xu2017video} and video retrieval on SSv2-Label~\cite{lei2022revealing}, SSv2-Template~\cite{lei2022revealing}. As our evaluation metrics, we report the averaged R@\{1,5,10\} on SSv2-* and R@1 on VidQA. Since VidQA needs a multimodal fusion (MF) encoder to generate the answers, we cannot report the results without the MF module (i.e., Columns 1,2 in Row 2 in Tab.~\ref{tab:ablation_more_datasets}). Our results indicate that our conclusions (i.e., the importance of temporal modeling, multimodal fusion, and joint image+video pre-training) also hold on these tasks/datasets.

\begin{table}
    \centering
    \small
    \begin{tabular}{lcccc}
    \toprule
       Masking Ratio & 15\% & 50\% & 75\% \\
       \midrule
       Accuracy & 69.2\% & \textbf{70.8\%} & 69.9\% \\
    \bottomrule
    \end{tabular}
    \caption{We study the masking ratio for the Masked Language Modeling (MLM) pretraining objective.  We experiment with 4 frames using our final model on the 5M corpus.}
    \label{tab:tips_enc}
\end{table}

\begin{table}
    \centering
    \small
    \setlength{\tabcolsep}{3pt}
    \begin{tabular}{lcccc}
    \toprule
         & Mean Pool. & + Temp. Attn & + MF & + Img Data \\
        \midrule
        SSv2-\{L,T\} & 72.3 & 80.2 & 81.3 & \textbf{82.7} \\
        M-QA & N/A & N/A & 42.7 & \textbf{43.6} \\
    \bottomrule
    \end{tabular}
    \caption{The analysis on more tasks/datasets. SSv2-L and SSv2-T refers to SSv2-Label and SSv2-Template datasets~\cite{lei2022revealing}. M-QA refers to MSRVTT-QA~\cite{xu2017video}.}
    \label{tab:ablation_more_datasets}
\end{table}

\subsection{Additional Temporal Modeling Baselines}
\label{appendix:temporal_modeling}
As discussed in the main draft, our first step is to extend our initial image transformer to video via a temporal modeling mechanism.
Such a temporal modeling mechanism would enable training our model on multiple frames for more robust VidL spatiotemporal representation learning. 
For this part of our empirical study, we experiment with the following temporal modeling schemes using 4-frame inputs and pretrained on WebVid-2M~\cite{bain2021frozen}. 
Besides the four temporal modeling baselines (i.e., mean pooling (MP), late temporal attention (L-TA), temporal convolution (TC), and temporal attention (TA)) that we included in the main draft, we further study Temporal Attention via Prompts (TA-P) and Window Attention (WA). We describe each of these baselines in more detail below: 

{
\begin{itemize}
\item  \textbf{Temporal Attention via Prompts (TA-P).} Following, several previous methods~\cite{xue2022clip,ni2022expanding} we implement a baseline that uses temporal attention via prompt tokens.
As shown in Figure~\ref{fig:temporal_modeling},
we first add $m$ prompt tokens to each frame. 
Then, these prompt tokens attend to each other via temporal attention~\cite{gberta_2021_ICML} to exchange frame-level information.
Finally, all frame-level image tokens and prompt tokens for that frame attend to each other via spatial attention.
Our TA-P scheme follows the same implementation as in~\cite{ni2022expanding}.

\item  \textbf{Window Attention (WA).} Similar to Swin~\cite{liu2021swin}, the spatial-temporal tokens are divided into cuboids of size $T\times k \times k$, where $T$ is the number of frames and $k$ is the window size. WA is performed inside each cuboid. Similar to Temporal Attention, the WA is inserted before the spatial attention as in~\cite{gberta_2021_ICML}. 
We experiment with $k=2$ and $k=7$. Larger $k$ leads to an out-of-memory error.

\end{itemize}
}

\noindent We also illustrate these attention mechanisms in Figure~\ref{fig:temporal_modeling}.
Furthermore, for completeness, below, we also describe the four baselines included in the main draft of the paper.
{
\begin{itemize}
\item \textbf{Mean Pooling (MP).} In this variant, the visual encoder processes input frames independently and averages their frame-wise scores for the video-level score as in~\cite{luo2022clip4clip}.
\item  \textbf{Late Temporal Attention (L-TA).} In this variant, we attach 2 Transformer layers to an image encoder, which then aggregates temporal information across all input frames. 
\item \textbf{Temporal Convolution (TC).} We insert a TC block before the spatial attention in each ViT layer.
The TC block consists of a linear down-projection layer with hidden size 384, a depth-wise $3\times1\times1$ convolution as in~\cite{Tran_2019_ICCV}, a ReLU activation, and a linear up-projection layer.
\item  \textbf{Temporal Attention (TA).} We insert a TA before spatial attention in each layer as in TimeSformer~\cite{gberta_2021_ICML}.
\end{itemize}
}

\noindent
As shown in Table~\ref{tab:temporal_modeling}, Temporal Attention outperforms Temporal Convolution and Temporal Attention via Prompts by 2.1\% and 6.8\% respectively on averaged top-\{1,5,10\} accuracy. Window Attention with window sizes of $k=2$ and $k=7$ outperforms Temporal Attention by 0.2\% and 0.7\% respectively. These results indicate that high temporal modeling capacity is important in VidL models. As Window Attention has $k\times$ the computational and memory cost and limited performance improvement compared with Temporal Attention, we choose Temporal Attention as our final temporal modeling blocks.

\begin{table}
    \centering
    \small
    \setlength{\tabcolsep}{4pt}
    \begin{tabular}{lccccc}
        \toprule
        Module & M & D & A & Avg. & Mem(GB) \\
        \midrule
        Mean Pooling & 49.4 & 53.7 & 46.4 & 50.1 & \textbf{9.3}\\
        Late Temp. Attn & 50.3 & 54.3 & 46.0 & 50.6 & 10.3 \\
        Temp. Conv & 53.0 & 58.2 & 52.7 & 54.6 & 10.3\\
        Temp. Attn & 53.7 & \textbf{60.9} & 55.6 & 56.7 & 11.4\\
        \midrule
        Temp. Attn Promp. & 49.5 & 52.7 & 46.6 & 49.9 & 10.3\\
        Wind. Attn ($k=2$) & \textbf{55.4} & 59.0 & 56.2 & 56.9 & 12.5\\
        Wind. Attn ($k=7$) & 54.6 & 59.9 & \textbf{57.7} & \textbf{57.4} & 18.1\\
        \bottomrule
    \end{tabular}
    \caption{We study various temporal modeling schemes. 
    M, D and A represents MSR-VTT, DiDeMo and ActivityNet-Captions. The accuracies are averaged R-\{1,5,10\}.
    GPU memory is measured with a batch size of 32 and gradient checkpointing enabled.
    Temporal Attention is the same as Window Attention with $k=1$. We observe that a larger temporal modeling capacity leads to higher performance. However, Window Attention with large window size (i.e., $k=7$) only has slight benefits (+0.6\%) compared to Temporal Attention but a large increased GPU memory consumption. Thus, we use temporal attention for our subsequent experiments due to a favorable computational cost and accuracy balance.  These experiments are conducted with 4-frame inputs, without a multimodal fusion encoder, and using the VTC loss as described in Step 1 of the main draft.}
    \label{tab:temporal_modeling}
\end{table}

{\small
\bibliographystyle{ieee_fullname}
\bibliography{egbib.bib}

\begin{thebibliography}{10}\itemsep=-1pt

\bibitem{anne2017localizing}
Lisa Anne~Hendricks, Oliver Wang, Eli Shechtman, Josef Sivic, Trevor Darrell,
  and Bryan Russell.
\newblock Localizing moments in video with natural language.
\newblock In {\em Proceedings of the IEEE international conference on computer
  vision}, pages 5803--5812, 2017.

\bibitem{bain2021frozen}
Max Bain, Arsha Nagrani, G{\"u}l Varol, and Andrew Zisserman.
\newblock Frozen in time: A joint video and image encoder for end-to-end
  retrieval.
\newblock In {\em Proceedings of the IEEE/CVF International Conference on
  Computer Vision}, pages 1728--1738, 2021.

\bibitem{bain2022clip}
Max Bain, Arsha Nagrani, G{\"u}l Varol, and Andrew Zisserman.
\newblock A clip-hitchhiker's guide to long video retrieval.
\newblock {\em arXiv preprint arXiv:2205.08508}, 2022.

\bibitem{bao2021beit}
Hangbo Bao, Li Dong, and Furu Wei.
\newblock Beit: Bert pre-training of image transformers.
\newblock {\em arXiv preprint arXiv:2106.08254}, 2021.

\bibitem{gberta_2021_ICML}
Gedas Bertasius, Heng Wang, and Lorenzo Torresani.
\newblock Is space-time attention all you need for video understanding?
\newblock In {\em Proceedings of the International Conference on Machine
  Learning (ICML)}, July 2021.

\bibitem{buch2022revisiting}
Shyamal Buch, Crist{\'o}bal Eyzaguirre, Adrien Gaidon, Jiajun Wu, Li Fei-Fei,
  and Juan~Carlos Niebles.
\newblock Revisiting the" video" in video-language understanding.
\newblock In {\em Proceedings of the IEEE/CVF Conference on Computer Vision and
  Pattern Recognition}, pages 2917--2927, 2022.

\bibitem{byun2022grit}
Jaeseok Byun, Taebaek Hwang, Jianlong Fu, and Taesup Moon.
\newblock Grit-vlp: Grouped mini-batch sampling for efficient vision and
  language pre-training.
\newblock In {\em European Conference on Computer Vision}, pages 395--412.
  Springer, 2022.

\bibitem{caba2015activitynet}
Fabian Caba~Heilbron, Victor Escorcia, Bernard Ghanem, and Juan Carlos~Niebles.
\newblock Activitynet: A large-scale video benchmark for human activity
  understanding.
\newblock In {\em Proceedings of the ieee conference on computer vision and
  pattern recognition}, pages 961--970, 2015.

\bibitem{changpinyo2021conceptual}
Soravit Changpinyo, Piyush Sharma, Nan Ding, and Radu Soricut.
\newblock Conceptual 12m: Pushing web-scale image-text pre-training to
  recognize long-tail visual concepts.
\newblock In {\em Proceedings of the IEEE/CVF Conference on Computer Vision and
  Pattern Recognition}, pages 3558--3568, 2021.

\bibitem{chen2016training}
Tianqi Chen, Bing Xu, Chiyuan Zhang, and Carlos Guestrin.
\newblock Training deep nets with sublinear memory cost.
\newblock {\em arXiv preprint arXiv:1604.06174}, 2016.

\bibitem{chen2015microsoft}
Xinlei Chen, Hao Fang, Tsung-Yi Lin, Ramakrishna Vedantam, Saurabh Gupta, Piotr
  Doll{\'a}r, and C~Lawrence Zitnick.
\newblock Microsoft coco captions: Data collection and evaluation server.
\newblock {\em arXiv preprint arXiv:1504.00325}, 2015.

\bibitem{chen2020uniter}
Yen-Chun Chen, Linjie Li, Licheng Yu, Ahmed El~Kholy, Faisal Ahmed, Zhe Gan, Yu
  Cheng, and Jingjing Liu.
\newblock Uniter: Universal image-text representation learning.
\newblock In {\em European conference on computer vision}, pages 104--120.
  Springer, 2020.

\bibitem{devlin2018bert}
Jacob Devlin, Ming-Wei Chang, Kenton Lee, and Kristina Toutanova.
\newblock Bert: Pre-training of deep bidirectional transformers for language
  understanding.
\newblock {\em arXiv preprint arXiv:1810.04805}, 2018.

\bibitem{dosovitskiy2020image}
Alexey Dosovitskiy, Lucas Beyer, Alexander Kolesnikov, Dirk Weissenborn,
  Xiaohua Zhai, Thomas Unterthiner, Mostafa Dehghani, Matthias Minderer, Georg
  Heigold, Sylvain Gelly, et~al.
\newblock An image is worth 16x16 words: Transformers for image recognition at
  scale.
\newblock {\em arXiv preprint arXiv:2010.11929}, 2020.

\bibitem{dou2022empirical}
Zi-Yi Dou, Yichong Xu, Zhe Gan, Jianfeng Wang, Shuohang Wang, Lijuan Wang,
  Chenguang Zhu, Pengchuan Zhang, Lu Yuan, Nanyun Peng, et~al.
\newblock An empirical study of training end-to-end vision-and-language
  transformers.
\newblock In {\em Proceedings of the IEEE/CVF Conference on Computer Vision and
  Pattern Recognition}, pages 18166--18176, 2022.

\bibitem{fan2021multiscale}
Haoqi Fan, Bo Xiong, Karttikeya Mangalam, Yanghao Li, Zhicheng Yan, Jitendra
  Malik, and Christoph Feichtenhofer.
\newblock Multiscale vision transformers.
\newblock In {\em Proceedings of the IEEE/CVF International Conference on
  Computer Vision}, pages 6824--6835, 2021.

\bibitem{feichtenhofer2020x3d}
Christoph Feichtenhofer.
\newblock X3d: Expanding architectures for efficient video recognition.
\newblock In {\em Proceedings of the IEEE/CVF Conference on Computer Vision and
  Pattern Recognition}, pages 203--213, 2020.

\bibitem{feichtenhofer2022masked}
Christoph Feichtenhofer, Haoqi Fan, Yanghao Li, and Kaiming He.
\newblock Masked autoencoders as spatiotemporal learners.
\newblock {\em arXiv preprint arXiv:2205.09113}, 2022.

\bibitem{fu2021violet}
Tsu-Jui Fu, Linjie Li, Zhe Gan, Kevin Lin, William~Yang Wang, Lijuan Wang, and
  Zicheng Liu.
\newblock Violet: End-to-end video-language transformers with masked
  visual-token modeling.
\newblock {\em arXiv preprint arXiv:2111.12681}, 2021.

\bibitem{fu2022empirical}
Tsu-Jui Fu, Linjie Li, Zhe Gan, Kevin Lin, William~Yang Wang, Lijuan Wang, and
  Zicheng Liu.
\newblock An empirical study of end-to-end video-language transformers with
  masked visual modeling.
\newblock {\em arXiv preprint arXiv:2209.01540}, 2022.

\bibitem{gao2021clip2tv}
Zijian Gao, Jingyu Liu, Sheng Chen, Dedan Chang, Hao Zhang, and Jinwei Yuan.
\newblock Clip2tv: An empirical study on transformer-based methods for
  video-text retrieval.
\newblock {\em arXiv preprint arXiv:2111.05610}, 2021.

\bibitem{goyal2017accurate}
Priya Goyal, Piotr Doll{\'a}r, Ross Girshick, Pieter Noordhuis, Lukasz
  Wesolowski, Aapo Kyrola, Andrew Tulloch, Yangqing Jia, and Kaiming He.
\newblock Accurate, large minibatch sgd: Training imagenet in 1 hour.
\newblock {\em arXiv preprint arXiv:1706.02677}, 2017.

\bibitem{goyal2017something}
Raghav Goyal, Samira Ebrahimi~Kahou, Vincent Michalski, Joanna Materzynska,
  Susanne Westphal, Heuna Kim, Valentin Haenel, Ingo Fruend, Peter Yianilos,
  Moritz Mueller-Freitag, et~al.
\newblock The" something something" video database for learning and evaluating
  visual common sense.
\newblock In {\em Proceedings of the IEEE international conference on computer
  vision}, pages 5842--5850, 2017.

\bibitem{he2022masked}
Kaiming He, Xinlei Chen, Saining Xie, Yanghao Li, Piotr Doll{\'a}r, and Ross
  Girshick.
\newblock Masked autoencoders are scalable vision learners.
\newblock In {\em Proceedings of the IEEE/CVF Conference on Computer Vision and
  Pattern Recognition}, pages 16000--16009, 2022.

\bibitem{hu2022scaling}
Xiaowei Hu, Zhe Gan, Jianfeng Wang, Zhengyuan Yang, Zicheng Liu, Yumao Lu, and
  Lijuan Wang.
\newblock Scaling up vision-language pre-training for image captioning.
\newblock In {\em Proceedings of the IEEE/CVF Conference on Computer Vision and
  Pattern Recognition}, pages 17980--17989, 2022.

\bibitem{iashin2020multi}
Vladimir Iashin and Esa Rahtu.
\newblock Multi-modal dense video captioning.
\newblock In {\em Proceedings of the IEEE/CVF Conference on Computer Vision and
  Pattern Recognition Workshops}, pages 958--959, 2020.

\bibitem{kay2017kinetics}
Will Kay, Joao Carreira, Karen Simonyan, Brian Zhang, Chloe Hillier, Sudheendra
  Vijayanarasimhan, Fabio Viola, Tim Green, Trevor Back, Paul Natsev, et~al.
\newblock The kinetics human action video dataset.
\newblock {\em arXiv preprint arXiv:1705.06950}, 2017.

\bibitem{kim2021vilt}
Wonjae Kim, Bokyung Son, and Ildoo Kim.
\newblock Vilt: Vision-and-language transformer without convolution or region
  supervision.
\newblock In {\em International Conference on Machine Learning}, pages
  5583--5594. PMLR, 2021.

\bibitem{krishna2017dense}
Ranjay Krishna, Kenji Hata, Frederic Ren, Li Fei-Fei, and Juan Carlos~Niebles.
\newblock Dense-captioning events in videos.
\newblock In {\em Proceedings of the IEEE international conference on computer
  vision}, pages 706--715, 2017.

\bibitem{krishna2017visual}
Ranjay Krishna, Yuke Zhu, Oliver Groth, Justin Johnson, Kenji Hata, Joshua
  Kravitz, Stephanie Chen, Yannis Kalantidis, Li-Jia Li, David~A Shamma, et~al.
\newblock Visual genome: Connecting language and vision using crowdsourced
  dense image annotations.
\newblock {\em International journal of computer vision}, 123(1):32--73, 2017.

\bibitem{lei2022revealing}
Jie Lei, Tamara~L Berg, and Mohit Bansal.
\newblock Revealing single frame bias for video-and-language learning.
\newblock {\em arXiv preprint arXiv:2206.03428}, 2022.

\bibitem{lei2021less}
Jie Lei, Linjie Li, Luowei Zhou, Zhe Gan, Tamara~L Berg, Mohit Bansal, and
  Jingjing Liu.
\newblock Less is more: Clipbert for video-and-language learning via sparse
  sampling.
\newblock In {\em Proceedings of the IEEE/CVF Conference on Computer Vision and
  Pattern Recognition}, pages 7331--7341, 2021.

\bibitem{lei2018tvqa}
Jie Lei, Licheng Yu, Mohit Bansal, and Tamara~L Berg.
\newblock Tvqa: Localized, compositional video question answering.
\newblock {\em arXiv preprint arXiv:1809.01696}, 2018.

\bibitem{li2022align}
Dongxu Li, Junnan Li, Hongdong Li, Juan~Carlos Niebles, and Steven~CH Hoi.
\newblock Align and prompt: Video-and-language pre-training with entity
  prompts.
\newblock In {\em Proceedings of the IEEE/CVF Conference on Computer Vision and
  Pattern Recognition}, pages 4953--4963, 2022.

\bibitem{li2022blip}
Junnan Li, Dongxu Li, Caiming Xiong, and Steven Hoi.
\newblock Blip: Bootstrapping language-image pre-training for unified
  vision-language understanding and generation.
\newblock {\em arXiv preprint arXiv:2201.12086}, 2022.

\bibitem{li2021align}
Junnan Li, Ramprasaath Selvaraju, Akhilesh Gotmare, Shafiq Joty, Caiming Xiong,
  and Steven Chu~Hong Hoi.
\newblock Align before fuse: Vision and language representation learning with
  momentum distillation.
\newblock {\em Advances in neural information processing systems},
  34:9694--9705, 2021.

\bibitem{li2020hero}
Linjie Li, Yen-Chun Chen, Yu Cheng, Zhe Gan, Licheng Yu, and Jingjing Liu.
\newblock Hero: Hierarchical encoder for video+ language omni-representation
  pre-training.
\newblock {\em arXiv preprint arXiv:2005.00200}, 2020.

\bibitem{li2022lavender}
Linjie Li, Zhe Gan, Kevin Lin, Chung-Ching Lin, Zicheng Liu, Ce Liu, and Lijuan
  Wang.
\newblock Lavender: Unifying video-language understanding as masked language
  modeling.
\newblock {\em arXiv preprint arXiv:2206.07160}, 2022.

\bibitem{lin2014microsoft}
Tsung-Yi Lin, Michael Maire, Serge Belongie, James Hays, Pietro Perona, Deva
  Ramanan, Piotr Doll{\'a}r, and C~Lawrence Zitnick.
\newblock Microsoft coco: Common objects in context.
\newblock In {\em European conference on computer vision}, pages 740--755.
  Springer, 2014.

\bibitem{lin2022eclipse}
Yan-Bo Lin, Jie Lei, Mohit Bansal, and Gedas Bertasius.
\newblock Eclipse: Efficient long-range video retrieval using sight and sound.
\newblock {\em arXiv preprint arXiv:2204.02874}, 2022.

\bibitem{liu2019use}
Yang Liu, Samuel Albanie, Arsha Nagrani, and Andrew Zisserman.
\newblock Use what you have: Video retrieval using representations from
  collaborative experts.
\newblock {\em arXiv preprint arXiv:1907.13487}, 2019.

\bibitem{liu2021swin}
Ze Liu, Yutong Lin, Yue Cao, Han Hu, Yixuan Wei, Zheng Zhang, Stephen Lin, and
  Baining Guo.
\newblock Swin transformer: Hierarchical vision transformer using shifted
  windows.
\newblock In {\em Proceedings of the IEEE/CVF International Conference on
  Computer Vision}, pages 10012--10022, 2021.

\bibitem{liu2022convnet}
Zhuang Liu, Hanzi Mao, Chao-Yuan Wu, Christoph Feichtenhofer, Trevor Darrell,
  and Saining Xie.
\newblock A convnet for the 2020s.
\newblock In {\em Proceedings of the IEEE/CVF Conference on Computer Vision and
  Pattern Recognition}, pages 11976--11986, 2022.

\bibitem{liu2022video}
Ze Liu, Jia Ning, Yue Cao, Yixuan Wei, Zheng Zhang, Stephen Lin, and Han Hu.
\newblock Video swin transformer.
\newblock In {\em Proceedings of the IEEE/CVF Conference on Computer Vision and
  Pattern Recognition}, pages 3202--3211, 2022.

\bibitem{loshchilov2016sgdr}
Ilya Loshchilov and Frank Hutter.
\newblock Sgdr: Stochastic gradient descent with warm restarts.
\newblock {\em arXiv preprint arXiv:1608.03983}, 2016.

\bibitem{loshchilov2017decoupled}
Ilya Loshchilov and Frank Hutter.
\newblock Decoupled weight decay regularization.
\newblock {\em arXiv preprint arXiv:1711.05101}, 2017.

\bibitem{lu2019vilbert}
Jiasen Lu, Dhruv Batra, Devi Parikh, and Stefan Lee.
\newblock Vilbert: Pretraining task-agnostic visiolinguistic representations
  for vision-and-language tasks.
\newblock {\em Advances in neural information processing systems}, 32, 2019.

\bibitem{luo2020univl}
Huaishao Luo, Lei Ji, Botian Shi, Haoyang Huang, Nan Duan, Tianrui Li, Jason
  Li, Taroon Bharti, and Ming Zhou.
\newblock Univl: A unified video and language pre-training model for multimodal
  understanding and generation.
\newblock {\em arXiv preprint arXiv:2002.06353}, 2020.

\bibitem{luo2022clip4clip}
Huaishao Luo, Lei Ji, Ming Zhong, Yang Chen, Wen Lei, Nan Duan, and Tianrui Li.
\newblock Clip4clip: An empirical study of clip for end to end video clip
  retrieval and captioning.
\newblock {\em Neurocomputing}, 508:293--304, 2022.

\bibitem{micikevicius2017mixed}
Paulius Micikevicius, Sharan Narang, Jonah Alben, Gregory Diamos, Erich Elsen,
  David Garcia, Boris Ginsburg, Michael Houston, Oleksii Kuchaiev, Ganesh
  Venkatesh, et~al.
\newblock Mixed precision training.
\newblock {\em arXiv preprint arXiv:1710.03740}, 2017.

\bibitem{miech2019howto100m}
Antoine Miech, Dimitri Zhukov, Jean-Baptiste Alayrac, Makarand Tapaswi, Ivan
  Laptev, and Josef Sivic.
\newblock Howto100m: Learning a text-video embedding by watching hundred
  million narrated video clips.
\newblock In {\em Proceedings of the IEEE/CVF International Conference on
  Computer Vision}, pages 2630--2640, 2019.

\bibitem{neimark2021video}
Daniel Neimark, Omri Bar, Maya Zohar, and Dotan Asselmann.
\newblock Video transformer network.
\newblock In {\em Proceedings of the IEEE/CVF International Conference on
  Computer Vision}, pages 3163--3172, 2021.

\bibitem{ni2022expanding}
Bolin Ni, Houwen Peng, Minghao Chen, Songyang Zhang, Gaofeng Meng, Jianlong Fu,
  Shiming Xiang, and Haibin Ling.
\newblock Expanding language-image pretrained models for general video
  recognition.
\newblock In {\em European Conference on Computer Vision}, pages 1--18.
  Springer, 2022.

\bibitem{ordonez2011im2text}
Vicente Ordonez, Girish Kulkarni, and Tamara Berg.
\newblock Im2text: Describing images using 1 million captioned photographs.
\newblock {\em Advances in neural information processing systems}, 24, 2011.

\bibitem{pan2022st}
Junting Pan, Ziyi Lin, Xiatian Zhu, Jing Shao, and Hongsheng Li.
\newblock St-adapter: Parameter-efficient image-to-video transfer learning for
  action recognition.
\newblock {\em arXiv preprint arXiv:2206.13559}, 2022.

\bibitem{paszke2019pytorch}
Adam Paszke, Sam Gross, Francisco Massa, Adam Lerer, James Bradbury, Gregory
  Chanan, Trevor Killeen, Zeming Lin, Natalia Gimelshein, Luca Antiga, et~al.
\newblock Pytorch: An imperative style, high-performance deep learning library.
\newblock {\em Advances in neural information processing systems}, 32, 2019.

\bibitem{radford2021learning}
Alec Radford, Jong~Wook Kim, Chris Hallacy, Aditya Ramesh, Gabriel Goh,
  Sandhini Agarwal, Girish Sastry, Amanda Askell, Pamela Mishkin, Jack Clark,
  et~al.
\newblock Learning transferable visual models from natural language
  supervision.
\newblock In {\em International Conference on Machine Learning}, pages
  8748--8763. PMLR, 2021.

\bibitem{ramesh2021zero}
Aditya Ramesh, Mikhail Pavlov, Gabriel Goh, Scott Gray, Chelsea Voss, Alec
  Radford, Mark Chen, and Ilya Sutskever.
\newblock Zero-shot text-to-image generation.
\newblock In {\em International Conference on Machine Learning}, pages
  8821--8831. PMLR, 2021.

\bibitem{seo2022end}
Paul~Hongsuck Seo, Arsha Nagrani, Anurag Arnab, and Cordelia Schmid.
\newblock End-to-end generative pretraining for multimodal video captioning.
\newblock In {\em Proceedings of the IEEE/CVF Conference on Computer Vision and
  Pattern Recognition}, pages 17959--17968, 2022.

\bibitem{sharma2018conceptual}
Piyush Sharma, Nan Ding, Sebastian Goodman, and Radu Soricut.
\newblock Conceptual captions: A cleaned, hypernymed, image alt-text dataset
  for automatic image captioning.
\newblock In {\em Proceedings of the 56th Annual Meeting of the Association for
  Computational Linguistics (Volume 1: Long Papers)}, pages 2556--2565, 2018.

\bibitem{singh2022flava}
Amanpreet Singh, Ronghang Hu, Vedanuj Goswami, Guillaume Couairon, Wojciech
  Galuba, Marcus Rohrbach, and Douwe Kiela.
\newblock Flava: A foundational language and vision alignment model.
\newblock In {\em Proceedings of the IEEE/CVF Conference on Computer Vision and
  Pattern Recognition}, pages 15638--15650, 2022.

\bibitem{sun2019videobert}
Chen Sun, Austin Myers, Carl Vondrick, Kevin Murphy, and Cordelia Schmid.
\newblock Videobert: A joint model for video and language representation
  learning.
\newblock In {\em Proceedings of the IEEE/CVF International Conference on
  Computer Vision}, pages 7464--7473, 2019.

\bibitem{tan2019lxmert}
Hao Tan and Mohit Bansal.
\newblock Lxmert: Learning cross-modality encoder representations from
  transformers.
\newblock {\em arXiv preprint arXiv:1908.07490}, 2019.

\bibitem{tong2022videomae}
Zhan Tong, Yibing Song, Jue Wang, and Limin Wang.
\newblock Videomae: Masked autoencoders are data-efficient learners for
  self-supervised video pre-training.
\newblock {\em arXiv preprint arXiv:2203.12602}, 2022.

\bibitem{Tran_2019_ICCV}
Du Tran, Heng Wang, Lorenzo Torresani, and Matt Feiszli.
\newblock Video classification with channel-separated convolutional networks.
\newblock In {\em Proceedings of the IEEE/CVF International Conference on
  Computer Vision (ICCV)}, October 2019.

\bibitem{vaswani2017attention}
Ashish Vaswani, Noam Shazeer, Niki Parmar, Jakob Uszkoreit, Llion Jones,
  Aidan~N Gomez, {\L}ukasz Kaiser, and Illia Polosukhin.
\newblock Attention is all you need.
\newblock {\em Advances in neural information processing systems}, 30, 2017.

\bibitem{wang2022all}
Alex~Jinpeng Wang, Yixiao Ge, Rui Yan, Yuying Ge, Xudong Lin, Guanyu Cai,
  Jianping Wu, Ying Shan, Xiaohu Qie, and Mike~Zheng Shou.
\newblock All in one: Exploring unified video-language pre-training.
\newblock {\em arXiv preprint arXiv:2203.07303}, 2022.

\bibitem{wang2018reconstruction}
Bairui Wang, Lin Ma, Wei Zhang, and Wei Liu.
\newblock Reconstruction network for video captioning.
\newblock In {\em Proceedings of the IEEE conference on computer vision and
  pattern recognition}, pages 7622--7631, 2018.

\bibitem{wang2022omnivl}
Junke Wang, Dongdong Chen, Zuxuan Wu, Chong Luo, Luowei Zhou, Yucheng Zhao,
  Yujia Xie, Ce Liu, Yu-Gang Jiang, and Lu Yuan.
\newblock Omnivl: One foundation model for image-language and video-language
  tasks.
\newblock {\em arXiv preprint arXiv:2209.07526}, 2022.

\bibitem{wang2022object}
Jinpeng Wang, Yixiao Ge, Guanyu Cai, Rui Yan, Xudong Lin, Ying Shan, Xiaohu
  Qie, and Mike~Zheng Shou.
\newblock Object-aware video-language pre-training for retrieval.
\newblock In {\em Proceedings of the IEEE/CVF Conference on Computer Vision and
  Pattern Recognition}, pages 3313--3322, 2022.

\bibitem{wang2021ufo}
Jianfeng Wang, Xiaowei Hu, Zhe Gan, Zhengyuan Yang, Xiyang Dai, Zicheng Liu,
  Yumao Lu, and Lijuan Wang.
\newblock Ufo: A unified transformer for vision-language representation
  learning.
\newblock {\em arXiv preprint arXiv:2111.10023}, 2021.

\bibitem{wang2022vlmixer}
Teng Wang, Wenhao Jiang, Zhichao Lu, Feng Zheng, Ran Cheng, Chengguo Yin, and
  Ping Luo.
\newblock Vlmixer: Unpaired vision-language pre-training via cross-modal
  cutmix.
\newblock In {\em International Conference on Machine Learning}, pages
  22680--22690. PMLR, 2022.

\bibitem{wang2022image}
Wenhui Wang, Hangbo Bao, Li Dong, Johan Bjorck, Zhiliang Peng, Qiang Liu, Kriti
  Aggarwal, Owais~Khan Mohammed, Saksham Singhal, Subhojit Som, et~al.
\newblock Image as a foreign language: Beit pretraining for all vision and
  vision-language tasks.
\newblock {\em arXiv preprint arXiv:2208.10442}, 2022.

\bibitem{xie2018rethinking}
Saining Xie, Chen Sun, Jonathan Huang, Zhuowen Tu, and Kevin Murphy.
\newblock Rethinking spatiotemporal feature learning: Speed-accuracy trade-offs
  in video classification.
\newblock In {\em Proceedings of the European conference on computer vision
  (ECCV)}, pages 305--321, 2018.

\bibitem{xu2017video}
Dejing Xu, Zhou Zhao, Jun Xiao, Fei Wu, Hanwang Zhang, Xiangnan He, and Yueting
  Zhuang.
\newblock Video question answering via gradually refined attention over
  appearance and motion.
\newblock In {\em Proceedings of the 25th ACM international conference on
  Multimedia}, pages 1645--1653, 2017.

\bibitem{xu2021videoclip}
Hu Xu, Gargi Ghosh, Po-Yao Huang, Dmytro Okhonko, Armen Aghajanyan, Florian
  Metze, Luke Zettlemoyer, and Christoph Feichtenhofer.
\newblock Videoclip: Contrastive pre-training for zero-shot video-text
  understanding.
\newblock {\em arXiv preprint arXiv:2109.14084}, 2021.

\bibitem{xu2016msr}
Jun Xu, Tao Mei, Ting Yao, and Yong Rui.
\newblock Msr-vtt: A large video description dataset for bridging video and
  language.
\newblock In {\em Proceedings of the IEEE conference on computer vision and
  pattern recognition}, pages 5288--5296, 2016.

\bibitem{xue2022clip}
Hongwei Xue, Yuchong Sun, Bei Liu, Jianlong Fu, Ruihua Song, Houqiang Li, and
  Jiebo Luo.
\newblock Clip-vip: Adapting pre-trained image-text model to video-language
  representation alignment.
\newblock {\em arXiv preprint arXiv:2209.06430}, 2022.

\bibitem{yang2021just}
Antoine Yang, Antoine Miech, Josef Sivic, Ivan Laptev, and Cordelia Schmid.
\newblock Just ask: Learning to answer questions from millions of narrated
  videos.
\newblock In {\em Proceedings of the IEEE/CVF International Conference on
  Computer Vision}, pages 1686--1697, 2021.

\bibitem{yang2022zero}
Antoine Yang, Antoine Miech, Josef Sivic, Ivan Laptev, and Cordelia Schmid.
\newblock Zero-shot video question answering via frozen bidirectional language
  models.
\newblock {\em arXiv preprint arXiv:2206.08155}, 2022.

\bibitem{yang2022unified}
Jianwei Yang, Chunyuan Li, Pengchuan Zhang, Bin Xiao, Ce Liu, Lu Yuan, and
  Jianfeng Gao.
\newblock Unified contrastive learning in image-text-label space.
\newblock In {\em Proceedings of the IEEE/CVF Conference on Computer Vision and
  Pattern Recognition}, pages 19163--19173, 2022.

\bibitem{yang2021causal}
Xu Yang, Hanwang Zhang, Guojun Qi, and Jianfei Cai.
\newblock Causal attention for vision-language tasks.
\newblock In {\em Proceedings of the IEEE/CVF Conference on Computer Vision and
  Pattern Recognition}, pages 9847--9857, 2021.

\bibitem{yu2022coca}
Jiahui Yu, Zirui Wang, Vijay Vasudevan, Legg Yeung, Mojtaba Seyedhosseini, and
  Yonghui Wu.
\newblock Coca: Contrastive captioners are image-text foundation models.
\newblock {\em arXiv preprint arXiv:2205.01917}, 2022.

\bibitem{yu2018joint}
Youngjae Yu, Jongseok Kim, and Gunhee Kim.
\newblock A joint sequence fusion model for video question answering and
  retrieval.
\newblock In {\em Proceedings of the European Conference on Computer Vision
  (ECCV)}, pages 471--487, 2018.

\bibitem{yu2019activitynet}
Zhou Yu, Dejing Xu, Jun Yu, Ting Yu, Zhou Zhao, Yueting Zhuang, and Dacheng
  Tao.
\newblock Activitynet-qa: A dataset for understanding complex web videos via
  question answering.
\newblock In {\em Proceedings of the AAAI Conference on Artificial
  Intelligence}, volume~33, pages 9127--9134, 2019.

\bibitem{yuan2021florence}
Lu Yuan, Dongdong Chen, Yi-Ling Chen, Noel Codella, Xiyang Dai, Jianfeng Gao,
  Houdong Hu, Xuedong Huang, Boxin Li, Chunyuan Li, et~al.
\newblock Florence: A new foundation model for computer vision.
\newblock {\em arXiv preprint arXiv:2111.11432}, 2021.

\bibitem{zellers2021merlot}
Rowan Zellers, Ximing Lu, Jack Hessel, Youngjae Yu, Jae~Sung Park, Jize Cao,
  Ali Farhadi, and Yejin Choi.
\newblock Merlot: Multimodal neural script knowledge models.
\newblock {\em Advances in Neural Information Processing Systems},
  34:23634--23651, 2021.

\bibitem{zeng2021multi}
Yan Zeng, Xinsong Zhang, and Hang Li.
\newblock Multi-grained vision language pre-training: Aligning texts with
  visual concepts.
\newblock {\em arXiv preprint arXiv:2111.08276}, 2021.

\bibitem{zhai2022lit}
Xiaohua Zhai, Xiao Wang, Basil Mustafa, Andreas Steiner, Daniel Keysers,
  Alexander Kolesnikov, and Lucas Beyer.
\newblock Lit: Zero-shot transfer with locked-image text tuning.
\newblock In {\em Proceedings of the IEEE/CVF Conference on Computer Vision and
  Pattern Recognition}, pages 18123--18133, 2022.

\bibitem{zhang2021vinvl}
Pengchuan Zhang, Xiujun Li, Xiaowei Hu, Jianwei Yang, Lei Zhang, Lijuan Wang,
  Yejin Choi, and Jianfeng Gao.
\newblock Vinvl: Revisiting visual representations in vision-language models.
\newblock In {\em Proceedings of the IEEE/CVF Conference on Computer Vision and
  Pattern Recognition}, pages 5579--5588, 2021.

\bibitem{zhou2020unified}
Luowei Zhou, Hamid Palangi, Lei Zhang, Houdong Hu, Jason Corso, and Jianfeng
  Gao.
\newblock Unified vision-language pre-training for image captioning and vqa.
\newblock In {\em Proceedings of the AAAI Conference on Artificial
  Intelligence}, volume~34, pages 13041--13049, 2020.

\bibitem{zhu2020actbert}
Linchao Zhu and Yi Yang.
\newblock Actbert: Learning global-local video-text representations.
\newblock In {\em Proceedings of the IEEE/CVF conference on computer vision and
  pattern recognition}, pages 8746--8755, 2020.

\bibitem{zolfaghari2018eco}
Mohammadreza Zolfaghari, Kamaljeet Singh, and Thomas Brox.
\newblock Eco: Efficient convolutional network for online video understanding.
\newblock In {\em Proceedings of the European conference on computer vision
  (ECCV)}, pages 695--712, 2018.

\end{thebibliography}
}

\end{document}